\documentclass[pdflatex,sn-apa]{sn-jnl}

\usepackage{graphicx}%
\usepackage{multirow}%
\usepackage{amsmath,amssymb,amsfonts}%
\usepackage{amsthm}%
\usepackage{mathrsfs}%
\usepackage[title]{appendix}%
\usepackage{xcolor}%
\usepackage{textcomp}%
\usepackage{manyfoot}%
\usepackage{booktabs}%
\usepackage{algorithm}%
\usepackage{algorithmicx}%
\usepackage{algpseudocode}%
\usepackage{listings}%

\theoremstyle{thmstyleone}%
\theoremstyle{thmstyletwo}%

\theoremstyle{thmstylethree}%

\begin{document}

\title[Dialogic Pedagogy for LLMs]{Dialogic Pedagogy for Large Language Models: Aligning Conversational AI with Proven Theories of Learning}

\author*[1]{\fnm{Russell} \sur{Beale}}\email{r.beale@bham.ac.uk}

\affil*[1]{\orgdiv{School of Computer Science}, \orgname{University of Birmimgham}, \orgaddress{\street{Edgbaston}, \city{Birmingham}, \postcode{B15 2TT}, \country{UK}}}


\abstract{Large Language Models (LLMs) are rapidly transforming education by enabling rich conversational learning experiences. This article provides a comprehensive review of how LLM-based conversational agents are being used in higher education, with extensions to secondary and lifelong learning contexts. We synthesize existing literature on LLMs in education and theories of conversational and dialogic pedagogy – including Vygotsky’s sociocultural learning (scaffolding and the Zone of Proximal Development), the Socratic method, and Laurillard’s conversational framework – and examine how prompting strategies and retrieval-augmented generation (RAG) can align LLM behaviors with these pedagogical theories, and how it can support personalized, adaptive learning. We map educational theories to LLM capabilities, highlighting where LLM-driven dialogue supports established learning principles and where it challenges or falls short of traditional pedagogical assumptions. Notable gaps in applying prior theories to LLMs are identified, such as the models’ tendency to provide direct answers instead of fostering co-construction of knowledge, and the need to account for the constant availability and broad but non-human expertise of LLM tutors. In response, we propose practical strategies to better align LLM interactions with sound pedagogy – for example, designing prompts that encourage Socratic questioning, scaffolded guidance, and student reflection, as well as integrating retrieval mechanisms to ensure accuracy and contextual relevance. Our aim is to bridge the gap between educational theory and the emerging practice of AI-driven conversational learning, offering insights and tools for making LLM-based dialogues more educationally productive and theory-aligned.}


\maketitle

\section{Introduction}

The emergence of powerful large language models such as GPT-3/4 and ChatGPT has introduced new possibilities for conversational learning in education. Educators and students can now interact with AI conversational agents that engage in dialogue, answer questions, and provide explanations on demand. These capabilities hint at a transformation in pedagogy: moving from conventional one-directional instruction or simple Q\&A tools to dynamic, dialogic pedagogy where learners and an AI tutor collaboratively construct understanding through conversation. Dialogue has long been considered central to learning – from Socratic questioning in ancient times to Vygotsky’s concept of knowledge co-construction through social interaction. However, applying these time-honored pedagogical approaches to artificial interlocutors raises new challenges and opportunities. LLMs are not human; they possess vast (but probabilistic) knowledge, are available 24/7, and tend to provide confident answers even when uncertain (though maybe that is a human trait...). 

This article explores how such characteristics are reshaping educational dialogue. We review recent literature on the use of LLMs and chatbots in education and revisit foundational theories of conversational learning. We then map these educational theories to LLM behaviors to illustrate points of alignment (e.g. how an LLM can scaffold a learner much like a human tutor) and misalignment (e.g. how an LLM might short-circuit the learning process by giving answers too readily). We also identify where existing pedagogical frameworks fall short in accounting for AI tutors’ unique attributes – for instance, prior theories did not envision an ever-available, non-human expert that students might consult at any time. Finally, we suggest concrete strategies to adapt LLM interactions for better educational outcomes, such as specialized prompting techniques and retrieval-augmented generation tweaks that enforce pedagogically desirable behaviors. The focus is on higher education, where autonomous learning and in-depth inquiry are paramount, but we also touch on secondary education and lifelong learning scenarios to the extent that findings generalize. By combining insights from peer-reviewed studies and theoretical analyses, this review aims to help educators and researchers harness LLMs in a way that is informed by and enriches established educational practice.

\section{Literature Review}

A growing body of research has examined the deployment of AI chatbots and LLM-based agents as teaching and learning tools, . 

\citet{thomas_conversational_1994} provides early empirical evidence that dialogue is a powerful pedagogical tool across age groups. Conversations between young children and adults strongly support intellectual development, and this “conversational learning” remains effective for older children as well – though traditional schooling often suppresses it. The paper points to the success of one-on-one dialogues (such as parent-child interactions, tutoring, and cooperative learning) in fostering understanding, highlighting the enduring value of conversational engagement in education.

\citet{zhang2024review} provide a comprehensive review of how chatbots (including AI-driven ones) have been used to support learning, explicitly focusing on pedagogical strategies and implementation contexts. They found that chatbots have been applied to support a wide range of learning activities – from drill and practice exercises and giving instructions, to participating in role-plays, facilitating collaborative tasks, assisting with writing, storytelling, educational games, and even open-ended debates. These applications span at least 14 different subject areas, and typically studies reported positive effects on learning outcomes and student attitudes (affective outcomes). From their analysis, the authors distilled a “RAISE” model for effective chatbot-assisted learning, which highlights five key design factors: Repetitiveness (the chatbot can repeatedly practice concepts with a student), Authenticity (the interaction or context is realistic or meaningful), Interactivity (high levels of two-way exchange and engagement), Student-Centeredness (the learner has agency and the chatbot adapts to learner needs), and Enjoyment (the experience is enjoyable, sustaining motivation). They also note that many learning theories can inform chatbot use. In fact, they list eight theoretical perspectives potentially relevant, including constructivism (students building knowledge, which chatbots can support by prompting learners), situated learning (contextualizing dialogue in real scenarios), multimedia learning (chatbots possibly using multiple modalities), self-regulated learning (chatbot as a coach for planning, monitoring learning), output hypothesis (having learners produce language or answers which the chatbot can elicit), flow theory (keeping students in an optimal challenge zone), collaborative learning (chatbot engaging in group-like interactions), and motivation theories. 

\citet{labadze_role_2023} also conducted a systematic review, looking at 67 studies on AI chatbots in education. They report that chatbots offer immediate, on-demand support for students (answering questions, explaining concepts, providing resources), functioning like virtual teaching assistants. Key benefits for students include help with homework and studying, personalized learning experiences, and skill development, while educators benefit from time saved (e.g. handling FAQs) and potential improvements in pedagogy (through additional support for students). However, the review also identifies major challenges: issues of reliability and accuracy of chatbot responses, and ethical concerns (e.g. data privacy, plagiarism, over-reliance). The authors stress that these challenges must be carefully managed for chatbots to be effective educational tools.

A conceptual framework for understanding AI conversational agents is provided by \citet{yusuf_pedagogical_2025}. By reviewing 92 sources, they categorize these agents along two primary dimensions: (1) Pedagogical application/purpose – including instructional (delivering content or tutoring), pastoral (supporting student welfare and motivation), and cognitive (facilitating thinking and reflection) purposes – further broken down by mode of study (in-class, online, etc.) and intent; and (2) Technological characteristics – such as whether the agent is embodied (a robot or avatar) or purely text-based, and what functional features it has (e.g. speech, adaptive feedback). Their framework helps map current applications and also highlights gaps. For example, many current agents focus on instruction, but there are opportunities to develop chatbots for assessment feedback, to prompt student reflection, or to assist with administrative tasks. The authors emphasize future research directions like improving personalization (tailoring interactions to individual learners) and increasing the media richness of interactions (possibly through multimodal interfaces or more natural dialogues). 

Focussing on the ideas of pedagogical application purpose, \citet{cao_ai_2023} present an experimental learning platform where students interact with multiple specialized chatbot personas, rather than just one. In a computer science course, they created four AI chatbot roles: an Instructor Bot (to deliver content or clarify concepts), a Peer Bot (to act as a learning companion or fellow student in discussions), a Career Advising Bot (to answer questions about career paths, linking coursework to professional development), and an Emotional Support Bot (to provide encouragement and address motivation or stress). These roles were deliberately chosen based on Self-Determination Theory (SDT) – a theory of motivation which says learners need to feel competent, autonomous, and related (socially connected). Each bot role maps onto these needs: the Instructor bot builds competence, the Peer bot and Emotional Supporter bot foster a sense of relatedness and supportive autonomy, and the Career bot connects learning to personal goals (supporting autonomous motivation). The system also encourages inquiry-based learning, meaning students are prompted to ask questions and explore problems with the bots rather than just receive information passively. In a month-long trial with 200 students, results (quantitative chat analyses and qualitative feedback) suggested that this multi-agent approach boosted student engagement and motivation compared to more traditional setups (like a single chatbot or even a human tutor in some comparisons). Students could benefit from different types of support on demand – for example, asking the Peer Bot for hints, then the Instructor Bot for a detailed explanation, getting reassurance from the Emotional Supporter, etc. This study demonstrates that carefully orchestrated multi-role LLMs can emulate a richer educational experience, addressing different facets of learning, from cognitive help to emotional and motivational support. It also illustrates a novel way to align AI behavior with educational psychology frameworks (here, SDT), yielding practical insights into how to design AI tutors that attend to students’ holistic needs.

Conversational agents are not new, however: \citet{bui_effects_2025} review pre-ChatGPT conversational agents (covering studies from 2004–2019) to understand what was known before the recent AI boom. They identified 23 relevant studies (29 chatbot implementations) in education. Despite being “pre-generative AI,” these chatbots (often rule-based or simpler AI) still showed positive effects on learning outcomes (generally medium to large effect sizes) and were received well by students in terms of usability. The implementations varied widely – different interface types (text vs. voice), different roles (tutor, peer, etc.), and domains. Notably, relatively few studies rigorously measured learning outcomes, and those that did often had small sample sizes or lacked randomization, pointing to methodological weaknesses in the literature. The authors argue that these earlier studies, despite limitations, demonstrate the potential of conversational agents and provide important context.  This helps ensure that current and future uses of LLM-based agents build on past lessons and are evaluated rigorously.

Focussing on the specifics of conversational approaches, \citet{chang2023prompting} explores how to explicitly incorporate the Socratic method into prompt design for LLMs. The Socratic method – characterized by asking guided questions to stimulate critical thinking rather than giving direct answers – is broken down into specific techniques (e.g. requesting definitions, using elenchus or probing questions to reveal contradictions, employing maieutics to draw out knowledge, and posing counterfactual or generalizing questions). Chang tested various prompt templates on GPT-3/ChatGPT and found that certain strategies produced not only accurate answers but also compelling justifications and creative elaborations. For instance, prompts that ask the model to answer a question by first questioning assumptions or definitions tended to yield more thorough reasoning in the responses. One notable finding is that if you explicitly state the dialogue’s goal and the user’s intent to the model at the outset (essentially “priming” the conversation with context about what the user is trying to achieve), the LLM’s subsequent responses better adhere to that intent and context. The paper demonstrates through examples that LLMs can simulate a Socratic tutor role – they can be prompted to ask the user a series of questions in return or to break down the problem – resulting in an interactive exchange that more closely resembles a human tutor guiding a student. Interestingly, Chang also remarks that some classic weaknesses of human Socratic teaching (such as students feeling intimidated or being afraid to respond due to power dynamics) might be mitigated with an AI tutor, since an LLM has no emotions or judgment – a point which suggests unique advantages of AI in applying this age-old pedagogic technique.

A theoretical perspective on using AI as an advanced tutor over a long term through the lens of Vygotsky’s sociocultural theory is offered by \citet{satra2025scaffolding}. Vygotsky introduced the concept of the Zone of Proximal Development (ZPD)\citep{vygotsky_mind_1978} – the range of tasks a learner can perform with guidance but not yet independently – and the idea of a “more knowledgeable other” who provides scaffolding within that zone. Sætra considers top human performers (“champions”) in domains like chess and Go, who traditionally struggle to find human mentors more skilled than themselves. He argues that AI, having surpassed human expertise in these domains, can act as a “more competent other” to even the best humans, pushing their skills further. The paper outlines a framework for AI-assisted expert learning: AI systems can provide scaffolding by giving feedback, insights, or challenges that lie just beyond the human champion’s current ability, thus keeping them in their ZPD for growth. Sætra notes that even AI systems not specifically built for education (for example, AlphaGo for playing Go) inadvertently perform teaching functions when humans learn from playing against them or analyzing their moves – illustrating that scaffolding can occur unintentionally. He further posits that the ideal scenario is not AI tutors in isolation, but a collaborative scaffold where AI, human peers, and human coaches all contribute to a learner’s development. This multi-source support acknowledges that AI can provide superhuman expertise and endless practice opportunities, while human mentors and peers contribute empathy, inspiration, and real-world contextualization. The work highlights potential limitations too – such as the risk of humans becoming overly reliant on AI – but overall suggests that with careful integration, AI tutors can extend Vygotskian theory into new territory: supporting learning at the very high end of expertise and throughout one’s lifelong learning journey.

Finally, \citet{sinha2025beyond} provides a critical and forward-looking commentary on the state of AI in Education (AIED), emphasizing that technology should not outpace pedagogy. The author argues that too often, AI tools are deployed in classrooms as fancy gadgets without a clear connection to sound learning theories – a situation described as “a solution in search of a problem.” Instead, Sinha calls for human-centered AIED, where AI is designed to augment human teaching and learning relationships, not replace them. Key points include the need for AI systems to support genuine relationship-building (e.g. fostering mentorship-like connections and trust, rather than just being question-answering machines) and rich scaffolding that challenges students appropriately (rather than simply automating tasks or giving answers that make learning too easy). The article suggests using real-world (“ecologically valid”) educational data – such as actual classroom interactions, multi-modal data of student engagement – to train AI, ensuring the technology is grounded in how students truly learn and behave, which can be messy and not always “rational” or uniform. Moreover, Sinha highlights that what counts as “good” use of AI in education can vary by discipline and pedagogical philosophy: for instance, a constructivist might use AI differently than a behaviorist. Therefore, interdisciplinary collaboration is needed; researchers with different theoretical orientations should work together to study AI in learning at scale, to ensure that these systems are informed by the full spectrum of learning sciences theory. This work identifies a gap in current practice – the lack of integration between AI design and educational theory – and urges that closing this gap is essential for AI to truly benefit learners.

In summary, the literature indicates that:

\begin{enumerate}
    \item Conversational agents (including LLM-based chatbots) can provide effective support for learning across many contexts, often improving engagement, personalization, and learning outcomes. Students benefit from immediate help and tailored instruction, and educators can offload some tasks to AI assistants.
    \item Challenges and limitations are consistently noted – particularly regarding the accuracy and reliability of LLM responses, ethical concerns (e.g. plagiarism, bias), and the need for pedagogical oversight (an AI is only as good as its design and the context in which it’s used). Many studies emphasize the importance of addressing these issues before AI tutors can be fully trusted in education.
    \item Researchers are actively trying to bridge AI capabilities with educational theory. For example, new frameworks and models link chatbot features to motivational psychology (as in Self-Determination Theory with multi-role bots) or to pedagogical principles (as in the RAISE model of Repetitiveness, Authenticity, Interactivity, Student-centeredness, and Enjoyment) . Several works explicitly call for grounding AI in proven theories of learning and instruction  – a theme that resonates strongly in recent meta-analyses and conceptual papers.

Overall, the literature suggests immense potential for LLMs to transform learning, but it also cautions that this potential will only be realized if we consciously design these tools in alignment with how people learn best.
\end{enumerate}

\section{Dialogic Pedagogy and Conversational Learning Theories}
The concept of learning through dialogue has rich roots in educational theory, and it provides a crucial lens for evaluating LLM-driven education. Whilst dialogic approaches tend to focus on a more structured approach where the characteristics of the dialogue are part of the learning process, conversational approaches tend to be more informal and naturalistic: both are relevant to LLMs.
 
\subsection{Socratic Method}
One of the oldest dialogic approaches, the Socratic method involves a teacher posing probing questions rather than giving answers, leading the student to deeper understanding. This method emphasizes critical thinking and illumination of ideas through back-and-forth inquiry. However, it also traditionally relies on a skilled teacher to pose the right questions and a dynamic interplay that can put students on the spot. In modern analysis, the Socratic approach is seen as a form of prompting strategy that can be implemented even with AI. As Chang (2023) showed, an LLM can be prompted to adopt Socratic questioning, using tactics like asking for definitions or counterexamples. Interestingly, some criticisms of human Socratic teaching (e.g. it may intimidate or frustrate students) might be less problematic with an AI tutor, which has no personal ego or judgment . Still, an AI might not intuitively know when to ask versus when to tell – this balance must be engineered through prompts or system design.

\subsection{Vygotsky’s Sociocultural Theory and ZPD}
Lev Vygotsky’s theory \citep{vygotsky_mind_1978} underscores that learning is socially mediated and that a more knowledgeable partner can help a learner perform just beyond their current independent ability. The Zone of Proximal Development (ZPD) is the sweet spot where this guided learning occurs. Traditionally, the “more knowledgeable other” is a parent, teacher, or peer who provides scaffolding – step-by-step support, hints, or encouragement – gradually removed as the learner grows competent. In the context of LLMs, one can ask: Can an AI serve as a more knowledgeable other, scaffolding a human learner?  \citet{satra2025scaffolding} explicitly addresses this, arguing that AIs, by virtue of sometimes exceeding human capabilities (for example, in specialized domains), can indeed operate in our ZPD and offer scaffolding support. For example, a coding student working on a problem might be unable to solve it alone (just outside their independent ability), but a chatbot like ChatGPT could provide a hint or a partial solution, enabling the student to complete the task – effectively scaffolding the student’s learning process. However, unlike humans, LLMs do not automatically know the learner’s actual level of understanding – they rely on what the student inputs. So while an AI can potentially play the tutor role, it may need mechanisms to gauge the student’s state (perhaps by asking the student to attempt a solution first or describe their reasoning). Vygotsky’s theory would encourage LLM usage that is contingent (adaptive to the learner’s responses) and fading (reducing help as learner improves). Current LLMs do not inherently do this, but with carefully structured interaction (or integration with student models), they could approximate it. It’s also worth noting that Vygotskyan dialogic learning is not just about a tutor’s help but also language as a tool for thought – encouraging students to articulate their thoughts. An LLM conversation that asks a student to “explain how you got your answer” can prompt the kind of self-talk and articulation that sociocultural theory values.

\subsection{Laurillard’s Conversational Framework}
\citet{laurillard_rethinking_2013} proposed a conversational model of learning specifically focusing on the dialogue between teacher and learner. In her framework, there is an iterative loop where the teacher’s conceptual knowledge and the student’s conceptual understanding interact through dialogue and action: the teacher articulates ideas, the learner expresses their understanding or misunderstanding, and both adjust – the teacher refines explanations or tasks, the learner revises their mental model. The framework emphasizes that learning is an interactive process of discussion, adaptation, reflection, and feedback. In an LLM-mediated scenario, we can imagine the AI fulfilling some teacher functions: presenting information, asking the student questions or for interpretations, and giving feedback on the student’s input: Tang et al. explore this approach in a Western Australian high school\citep{tang_dialogic_2024}. Some aspects of Laurillard’s framework have already been instantiated in intelligent tutoring systems (for example, systems that present a problem, ask the student to attempt it, then give feedback and additional explanation, etc.). An LLM can potentially make such interactions more fluid and natural-language based. However, Laurillard also underscores the importance of the student’s action in the real world and the teacher’s observation of it, especially for practical or skills learning. LLMs are limited to the dialogue itself and whatever the student describes – they can’t directly observe a student’s lab experiment or presentation outside the chat (unless coupled with other technology). So while LLMs can implement the conversational exchange part of the framework (conceptual discussions), they might not fully close the loop that involves empirical activity or experiential feedback, unless the educational platform integrates those aspects (for instance, having the student report on their activity for the LLM to discuss). Still, aligning LLM prompts with Laurillard’s model could mean ensuring that every interaction includes bidirectional communication: not just the student asking and AI telling, but the AI also asking the student to articulate understanding or do something with the knowledge, and then responding to that.

\subsection{Dialogic Teaching and Collaborative Inquiry}
Educationalists like Robin Alexander \citep{alexander2008culture} and researchers in dialogic pedagogy (e.g., \citet{wegerif2013dialogic}, \citet{mercer_explaining_2012, mercer_dialogic_2009}) have argued for classroom dialogue that is open, reciprocal, and cumulative. Mercer also talks about "Exploratory talk", where participants build on each other ideas, but rather than with cumulative talk which builds positively but uncritically of the contributions of others, he allows for statements and suggests to be offered for mutual consideration  which can be challenged and counter-challenged when justified, with alterative hypotheses given. "All actively participate and opinions are sought and considered before decisions are jointly made.  In exploratory scenarios, knowledge is made more publicly accountable and reasoning is more visible in the talk." \citep{mercer_sociocultural_2004}.

Dialogic teaching involves students building on each other’s ideas, asking questions, and engaging in reasoning together \citep{lee_can_2024, han_impact_2010, kuttal_trade-offs_2021}. In a one-on-one student-LLM interaction, we don’t have multiple students, but we can attempt to simulate some aspects of dialogic discourse. For instance, an LLM can be prompted to not only answer but also prompt the student back (“What do you think about this result?” or “Can you provide an example?”), encouraging the student to contribute ideas. This can also happen when the LLM is one voice in many, and can act purely constructively or in an exploratory manner.  Projects like the multi-role chatbots \citep{cao_ai_2023} attempt to simulate a community of inquiry with different agent roles, essentially giving a single learner exposure to multiple viewpoints or types of interaction (peer discussion vs. teacher guidance). This means that LLMs could orchestrate a dialogic space even if only one human is present, by embodying different voices. However, true dialogic pedagogy values the unpredictability and authenticity of multiple human contributions – an AI, if not carefully checked, could end up essentially dialoguing with itself in different guises, which might not produce the rich clash of ideas that genuine student-peer discussion would. That said, for a lone learner (say, a remote student or self-learner), an LLM might provide a form of simulated dialogue to think through issues. It could also introduce perspectives (e.g., “One could argue X, but another viewpoint is Y; which do you find more convincing?”) to push the learner to evaluate different arguments, mirroring techniques used in argumentative dialogue education. Importantly, studies into pair programming, which have dialogue-style interactions, show that women rate their self-confidence lower than men \citep{jarratt_large-scale_2019} and feel more insecure \citep{ying_their_2019};working with an AI agent that can be appropriately gendered and is non-threatening is therefore a positive boost to supporting female progression.

\subsection{Constructivism and Inquiry-Based Learning}
The constructivist view (Piagetian \citep{piaget_part_1964} and beyond) holds that learners construct knowledge best through active engagement and exploration. Dialogues that are led by the learner’s curiosity (inquiry-based) or that involve the learner solving problems with guidance align with constructivist pedagogy. LLMs can support this by allowing learners to drive the conversation – the student can pose questions, ask “why?” as many times as needed, or request clarification until they build understanding. Unlike a static textbook, an LLM can adapt to the learner’s line of inquiry. Conversely, if misused, an LLM could undermine constructivism by giving answers too easily, preventing the productive struggle or exploration that is crucial for deep learning. So aligning with this theory means using the LLM more as a guide on the side than a sage on the stage: prompting it to prompt the learner, to give hints or partial info, and let the learner make connections. The inquiry-based paradigm mentioned in \citet{cao_ai_2023} is an example where students are encouraged to ask questions and the AI supports that process rather than simply delivering declarative content.

\subsection{Self-Regulated Learning (SRL)}
In lifelong learning contexts especially, the ability of learners to plan, monitor, and reflect on their own learning is key. Conversational agents can potentially act as meta-cognitive coaches – e.g., asking a student to set goals, or prompting them to reflect on what they have learned. Some of the theories identified by \citet{zhang2024review} – like self-regulated learning theory \citep{zimmerman2002becoming} and motivation theories – speak to this. An LLM could be instructed to periodically ask, “Shall we recap what we’ve covered so far?” or “How confident do you feel about this concept?” This mimics strategies a human tutor might use to cultivate self-awareness in the learner. Additionally, an AI’s constant availability means a motivated learner can structure their own learning schedule with the AI’s help (it’s always there to quiz them or explain things at 2am), which is a boon for self-directed learning. In secondary education, teachers could use an LLM to support students outside class hours – effectively giving them a tool to regulate their own practice and inquiry when a teacher isn’t present.

In mapping these theories to LLM behaviors, we observe areas of alignment as well as mismatches. Alignments include the LLM’s capacity to provide immediate responsive dialogue, supporting ideas from dialogic and conversational frameworks, its ability to tailor responses to student questions supporting personalized scaffolding in the ZPD when done carefully, and even to emulate certain roles such as Socratic questioner, peer, etc. that have known pedagogical value. Mismatches (which we detail further in the next section on gaps) include the lack of true social presence or emotional understanding in an LLM, affecting the relational aspect of pedagogy, the risk of oversimplifying a task by directly solving it, contrary to constructivist ideals of struggle and discovery, and the challenge of ensuring the AI’s guidance is always pedagogically sound: human tutors have wisdom and context that an LLM might lack without being explicitly programmed for it. Nonetheless, with intentional design, many educational theories can be at least partially instantiated in LLM-driven learning activities. For higher education students, who often engage in self-driven inquiry and dialogue in seminars, an LLM can act as a surrogate discussion partner – not replacing peer or professor interaction, but supplementing it. For secondary students, an LLM could serve as a personal tutor that reinforces the day’s lessons or helps with homework through guided conversation, ideally following the teacher’s pedagogical approach. For lifelong learners, LLMs provide a conversational interface to the world’s knowledge that can adapt to their goals – essentially a personal mentor available on-demand. Each of these use cases can be enhanced by consciously aligning the interaction patterns with proven educational practices - for example, an adult learner might prefer an andragogical approach where the LLM acts more like a consultant or coach rather than drilling basics \citep{knowles_modern_1980}.

\section{Limitations of Applying Traditional Pedagogy to LLMs}

While the promise of LLMs in enacting dialogic and scaffolded learning is evident, it’s equally important to acknowledge where traditional pedagogical theories struggle to account for the realities of AI tutors, and conversely, where AI-driven practices challenge the assumptions of those theories. Some key gaps and limitations include:

\subsection{Over-Directness vs. Productive Struggle}
Many pedagogical approaches (Socratic, constructivist, etc.) deliberately avoid simply handing the student the answer. Instead, they engage the learner in a process of thinking (through hints, questions, etc.), because struggling with a problem is part of learning. Out-of-the-box LLM behavior, however, tends to be very answer-oriented – if a student asks a question, the default is for the AI to provide a direct answer or explanation. This is in part due to their training (models like GPT are rewarded for providing a correct and complete answer to the user’s query) and the typical user expectations of AI assistants. From a theoretical view, this is a limitation: the LLM might short-circuit the learning process by doing the work for the student. Prior pedagogical theories did not envision a scenario where the “more knowledgeable other” would instantly solve problems; rather, even a good human tutor paces the assistance. There’s a deficiency in our pedagogical models here – they assume the tutor has the discipline to hold back the full solution. An AI does not inherently have that, unless instructed. Thus, if we naively apply LLMs, we violate the spirit of guided discovery. Students may become passive recipients of knowledge again (a regression to instructionism) but with a false sense of accomplishment since the AI did the heavy lifting.

\subsection{Lack of Assessment of Learner Understanding}
Traditional tutoring involves a lot of diagnostics – a tutor infers from a student’s responses what they have misunderstood or where their skill level lies, and then adjusts the instruction.  LLMs currently have no genuine model of the student’s knowledge state; they only have the immediate context of the conversation (and whatever the student explicitly says or whatever the system is programmed to track). This means an AI might overshoot or undershoot the ZPD. For example, a student asks a calculus question; if the AI explanation uses terms or steps too advanced (assuming the student knows more than they do) or conversely, explains in a patronizing, too-basic way, the scaffolding fails. Vygotsky’s theory assumes the tutor can roughly sense the ZPD through interaction, but an LLM is not equipped with theory-of-mind or student modeling out of the box. Prior theories do not provide guidance on how to handle a tutor that doesn’t naturally perceive the student’s state – that’s something the design of the AI system must supply (perhaps via quizzes, or by the student explicitly telling the AI their background). This is a gap because if we drop LLMs into educational settings without that adaptation, they might either bore students or confuse them, whereas a human would calibrate.

Developments in this area are rapid, however.  For example, ChatGPT 3o-mini-high remembers context across a conversational thread, but each new thread is a clean slate, and it knows nothing about who it is interacting with or what has gone before.  ChatGPT4.0 also remembers context within the conversation:
\begin{quote}
{\fontfamily{cmss}\selectfont
\small
    Yep — \textbf{within a single conversation}, I retain \textbf{the full context}, including all steps you've taken, what you've understood, what you're still working through, and anything you've said along the way. So if you're halfway through figuring out, say, how backpropagation works in neural networks or how subscription logic works in a web app, I can pick up right from where you left off — \textit{in this conversation}.

You don’t need to re-explain or recap anything unless we switch topics dramatically or start a new chat.\\
}
\end{quote}

However, it also has a memory for across-chat conversations. This is not a transcript of the chat, but a record of key information:

\begin{quote}
{\fontfamily{cmss}\selectfont
\small
When memory is enabled, I can remember helpful information you share across conversations — so I can give you more consistent, personalized help. This isn’t the same as remembering everything like a transcript — it’s more like keeping notes on key facts, goals, and preferences.

What I can remember:\\
Your name or how you like to be addressed\\
Your projects, goals, or ongoing tasks (e.g., "building a GPT analysing tool web app")\\
Your preferences (e.g., prefer Python, like detailed explanations, want bullet points)\\
Your learning progress, if you tell me to remember it (e.g., "I understand supervised learning but still get stuck on clustering").

What I don’t remember by default:\\
Most everyday conversation unless it’s something you asked me to remember\\
Sensitive or personal info unless you ask me to store it\\
Anything from a past chat unless it was saved to memory

You control memory:\\
You can always ask me to:\\
Remember something: "Remember that I’m studying for the GRE and I’ve finished the Quant section."\\
Forget something: "Forget that I’m working on the AVE project."\\
See what I remember: Just say: "What do you remember about me?"\\
}
\end{quote}

Thus the latest LLMs have the inherent facility to remember learner progress across sessions and across conversations, if explicitly promoted to do so.  What is of key importance for educational technologists is that bespoke LLMs can be created that, \textit{by default}, will store this progress across sessions, so giving a personalised and continuously updated model of the user.

\begin{quote}
{\fontfamily{cmss}\selectfont
\small
What custom GPTs can do:\\
They share memory with your main ChatGPT (if memory is on), so they can remember your progress if you save it explicitly or set up the custom GPT to do so automatically.\\
You can instruct your custom GPT (in the "instructions" section when building it) to ask about progress, log what you've learned, or save milestones in your learning — so it feels like it remembers everything naturally.

What custom GPTs cannot do:\\
They can’t automatically remember every message by default unless memory is enabled and used intentionally.\\
They don’t bypass OpenAI’s memory rules, meaning sensitive data still won’t be remembered unless it's appropriate and asked for.\\
}
\end{quote}
(All quotes from ChatGPT 4.0 14th April 2025.)

The capability for complex user modelling is therefore available within the current release of LLMs.  Note too that they can build this model based not only on the formalised structures of the specific learning tasks or classroom environment, but also from the ongoing projects, tasks and outside interests that a learner may engage ChatGPTs help with as well, giving rise to a potentially rich and powerful conception of the user, their personality likes, dislikes, skills and challenges.

\subsection{Continuous Availability and Learner Dependence}
An LLM tutor is available anytime, which is fantastic for access – a learner can get help at 1 AM when no human tutor is available. However, this also means a student might lean on the AI for every small roadblock instead of attempting to think it through. Classical theories did not anticipate a helper that is literally always present. Concepts like productive failure or desirable difficulty in learning (where struggling and sometimes failing to retrieve an answer is beneficial to memory) could be undermined if the student always asks the AI at the first sign of difficulty. Essentially, self-regulation skills might suffer if learners are not guided in how and when to use the AI. The notion of “scaffolding” includes eventually removing the scaffold – but if the AI is always there, will the student ever operate without support? This is a new question: pedagogical models must evolve to consider how constant AI assistance interacts with learning autonomy. It might be that new norms or strategies are needed to ensure students don’t become overly dependent (for instance, deliberately designing the AI to sometimes refuse immediate answers and encourage the student to try first – akin to a human teacher saying “I won’t tell you that yet, see if you can figure it out”).

\subsection{Accuracy and Truthfulness}
Traditional pedagogy usually assumes the teacher or tutor provides correct and vetted information (and if they don’t, it’s a grave issue). LLMs, however, are prone to hallucinations – they can generate plausible-sounding but incorrect information. None of the classical theories of learning dealt with an interlocutor that might inadvertently lie or make things up. This is a critical limitation: a Socratic dialogue is pointless if the “tutor” asks misleading questions based on false premises, or a scaffold becomes harmful if it’s built on incorrect examples. Students (especially younger or less knowledgeable ones) might not detect the errors and could learn incorrect facts or methods. This calls into question the reliability of the AI as a conversational partner in learning. The literature (e.g., \citet{labadze_role_2023}) flags this as a top concern. We need solutions like retrieval augmentation, verification steps, or hybrid human-AI oversight to ensure accuracy, which we discuss in the next section. But the key point is, prior pedagogic frameworks have no built-in mechanism for an untrusted tutor. The assumption was the tutor knows the material. With AI, the role of the human teacher might shift towards monitoring the AI’s output for correctness, a new kind of “meta-tutoring” role. This dynamic isn’t covered by classical dialogic models.

\subsection{Social and Emotional Limitations}
A significant part of education, especially dialogic pedagogy, involves the emotional and relational dimension – trust, encouragement, responsiveness to frustration, etc. LLMs can emulate some sympathetic responses (e.g., saying “I understand this is tricky, let’s try together”), but they do not genuinely feel or perceive the student’s emotions. They also lack authentic human relationship qualities. For some learners, especially younger ones, the motivation to engage in a dialogue is tied to the relationship with the teacher or peers. An AI might not provide that same motivation or might feel impersonal after a while. Sinha (2025) emphasizes keeping AIED human-centered and not losing the personal connection. If educational practice leans too heavily on impersonal AI tutors, theories of motivation (like SDT, which highlights relatedness) suggest students might disengage or not develop certain social aspects of learning. Additionally, while an AI tutor won’t get angry or tired, it also doesn’t celebrate a student’s success in a heartfelt way or truly care about them, which could limit how inspiring or validating the learning experience is. This is a subtle limitation – hard to measure, but pedagogues are aware of how important the teacher’s enthusiasm or a peer’s genuine praise can be. Current AI can’t replicate that authenticity of emotion.

\subsection{Dialogic Depth and Authenticity}
Dialogic pedagogy thrives on authentic questions and the co-construction of ideas. A risk with AI is that dialogues become artificially structured or one-sided. For example, an LLM might ask a question because it was prompted to, but the question might not stem from a genuine “wonder” or confusion – sometimes human students ask questions they genuinely have, which leads to explorations neither the teacher nor students predicted. An AI is unlikely to have genuine curiosity (it knows or can generate the answer to its own questions). Thus, certain emergent properties of human dialogue – like a sudden insightful question that changes the lesson trajectory – might not occur with AI (unless the AI is instructed to simulate the ignorance of a student, which flips the roles). In essence, AI might make the dialogue too efficient and goal-directed, whereas human-human dialogue can meander and discover new things serendipitously. Our theories value that emergent dialogue, but with AI we might inadvertently stifle it by keeping conversations very QA structured.

\subsection{Prior Theoretical Gaps in Scale and Consistency}
Some pedagogical concepts like one-on-one tutoring (e.g., Bloom’s 2-sigma finding \citep{bloom_2_1984} that one-on-one tutoring is extremely effective) were historically hard to implement at scale. LLMs suddenly offer a way to give every student a personal tutor. This is wonderful, but it also exposes that our theories about tutoring were not tested for when every student gets unlimited tutoring. Would Bloom’s result hold if the tutor were an AI and the student could use it unlimitedly? Possibly yes for achievement, but what about other outcomes like sociability, creativity or independence? Also, the constant availability could raise issues of burnout or superficial engagement – if a student knows the AI will always explain, maybe they skim lectures expecting to fill gaps via chatbot later. Such shifts in student behavior are not accounted for in traditional pedagogy. We may need updated frameworks that assume AI support as a given, and thus focus on meta-skills: how to teach students to learn with an AI effectively (a bit like how information literacy became crucial once internet search was ubiquitous).

\subsection{Ethical and Cognitive Development Considerations}
Dialogic pedagogy, especially from thinkers like Paulo Freire \citep{freire_pedagogy_2000}, is tied to critical consciousness and the development of the learner as an independent thinker who can question and shape the dialogue. If an LLM inadvertently has certain biases (from its training data) and subtly steers conversations, it could influence learners’ thinking patterns without the learners’ awareness. For example, if every explanation subtly reflects a Silicon Valley techno-optimist worldview (because of the data it saw), students might not be exposed to alternative perspectives unless explicitly prompted. Traditional pedagogy relies on the diversity of human teachers and peers to bring multiple perspectives. If one AI model dominates educational dialogue, there is a risk of monologism creeping into what should be a dialogic process. Ensuring multiple viewpoints and critical thinking in AI-guided learning is an open challenge; older theories don’t account for the homogeneity of an AI’s voice.

There is also an ethical dimension to consider in terms of their use at all: training and running an LLM uses a significant amount of power - GPT4 took just under 29,000 MWh of electricity and emitted nearly 7,000 metric tons of CO\textsubscript{2} \citep{luccioni_power_2024}; querying them uses a fraction of this but as usage scales up so does the environmental impact: we should not ignore the cost that these systems are imposing on the world.

In light of these limitations, it’s clear that simply plugging LLMs into education is not a magic fix – we must adapt either the technology or our pedagogical strategies (likely both) to mitigate these issues. Researchers like \citet{sinha2025beyond} argue strongly that we need to keep pedagogy in the driver’s seat. The gaps identified above serve as a checklist of what to be mindful of: balancing answer-giving with questioning (to avoid over-directness), developing methods for the AI to gauge student understanding (perhaps via formative assessments or student self-reports), guiding students in healthy usage patterns of AI, implementing verification and retrieval to ensure accuracy, maintaining a human touch in learning (possibly by coupling AI with human mentors in a hybrid model), and fostering authentic, multi-perspective dialogue even in AI-mediated environments. The next section will suggest practical methods that have been proposed or can be devised to address many of these points.

\section{The Implications of Ubiquity}
One of the significant implications for pedagogical consideration is the likely ubiquity of LLMs in interaction and general life.  With smartphones, it is not their ability to make phone calls away from a particular tethered base station that defines them: instead it is their integration into every facet of everyday living, giving immediate access to social connections, internet searches, video feeds and every other digital convenience and distraction.  It is highly likely that LLMs will also find significant uptake and appear across all sorts of scenarios. Many people are moving towards using them as their oracle instead of Google or Bing \citep{chris_rowlands_goodbye_2025}, and there are colloquial stories of schoolchildren asking it whether they should eat their sandwiches or crisps first.  But because they are never bored, never think a question is stupid, never criticise, and can provide the answers to many things quickly, they are a perfect companion to the challenge of growing up and developing.  It is therefore almost certain that LLMs will be integrated into many other digital experiences and products, and widely and simply accessible on smartphones and laptops all the time: omnipresent and omnipotent.  This is unlike any other educational technology we have seen before: user models didn't sit in the living room with us; instructional guides on programming were not with us on the bus as well.  We must therefore consider that a pedagogy of LLM use in education is actually a pedagogy of interaction all the time, everywhere: it can be tailored to do specific things in specific situations, but exists in an ecosystem where other LLM interactions happen all the time as well, setting expectations in learners that we have not had to content with before. We are not simply talking about how to we get them to best support our learning - we are asking how to get them best support our lives.  This makes this new situation very different to the scenarios we have previously encountered.  

Widespread use of  LLMs provides us with some interesting pedagogical insights.  LLMs have been used in many programming courses to help develop skills (e.g. \citet{becker_programming_2023, ma_how_2024, bird_taking_2023, manfredi_mixed_2023} and whilst they generally find very positive effects, it is clear that students become wary of the errors LLMs typically make in not understanding some programming context properly; in other contexts they realise that references provided are hallucinated and don't exist despite being very believable.  Indeed, some research deliberately builds on the hallucinogenic properties of AI to better support creative writing \citep{blythe_artificial_2023}.   Even in cases where the LLMs are prompted to provide the correct answers (such as CoPilot when programming) students learn by seeing correct examples of syntax and by using the 'explain' facility of the model to get insights into why things are done as they are.  In other words, even without appropriate pedagogical design, experiencing much more rapid turnaround of code and have a conversation about it with the LLM to get commentary and feedback allows a learner to quickly build up experience in seeing how code is created and extended and revised, and to infer structure, syntax and algorithmic design by example.  These scenarios demonstrate that simply having the LLM present and available and responsive can provide the 1:1 support and engagement that students need to help develop their own learning.

The ubiquity also extends to accessibility benefits:  newer LLMs are excellent at interpreting normal speech and responding verbally in a natural manner.  Whilst not perfect - they are particularly poor at distinguishing between a pause in speaking and finishing, for example - they open up their power to people beyond the keyboard just as well.  Whilst digital access is certainly not ubiquitous, whether we consider it worldwide or within countries, and hence a problem it is somewhat mitigated that powerful computers to access the models are not needed - a laptop or, perhaps more usefully, just a, smartphone is all that is needed to access their capabilities. They also provide for supportive, permanently accessible educational partners, something that women may find easier to relate to: \citet{ying_their_2019} found this for pair programming, for example, and approaches that are more supportive of women (especially in technical topics where they are under-represented) is to be welcomed.

\section{Pedagogic Adaptation Strategies for LLM-Based Conversations}

To harness LLMs for truly effective conversational learning, we need to intentionally design how these AI agents interact with students. Based on the literature and theoretical considerations above, here are several practical strategies – including specific prompting techniques and system adjustments (like RAG) – to make LLM-student dialogues more pedagogically productive and aligned with best practices.

\subsection{Socratic Prompting Templates}
Drawing on Chang (2023) and the classic Socratic method, instructors or developers can create prompt frameworks that encourage the LLM to teach by asking questions. For example, instead of a student query yielding a direct answer, the LLM could be instructed to respond with a guiding question or a hint. A prompt template might be: “If the user’s question is X, do not give the final answer immediately. First, ask a clarifying question or a simpler related question that leads them to think about X. Only provide the answer after they attempt to respond.” This effectively makes the LLM behave like a Socratic tutor. Concretely, if a student asks, “Why is the sky blue?”, the LLM might reply, “That’s a great question. Let me ask: what do you know about how light interacts with the atmosphere?” – initiating a dialogue. Such templates can incorporate specific Socratic techniques: elenchus (asking the student to examine inconsistencies, in particular to refute arguments: “Do you think X could be true given Y?”), maieutics (drawing out prior knowledge to make it clear and apparent: “Remember when we discussed reflection – how might that relate here?”), and counterfactuals (relating with what  has not happened or is not the case: “What if the atmosphere had no particles, what color do you think the sky would be?”). Implementing this requires prompt engineering or fine-tuning the model on example dialogues. It aligns with critical thinking goals and keeps the student mentally active. However, it is important to calibrate this – too many questions without any answers can frustrate learners. One practical approach is a “tiered Socratic” approach: the first time a concept appears, the AI asks the student to attempt it; if the student is stuck or answers incorrectly, the AI then gives a bigger hint or partial answer; if the student is still stuck, the AI finally explains in full. This mirrors human tutors who adjust their Socratic intensity based on student responses.

\subsection{Zone of Proximal Development (ZPD) Scaffolding}
We can design scaffolding prompts so that the LLM provides graduated assistance. For instance, when a student faces a complex problem, the LLM can break it down into sub-tasks or ask the student to attempt a solution step-by-step, offering help at each step. A strategy here is prompt chaining: the system might have an internal chain where it first says, “Let’s tackle this step by step. First, how would you approach…?” If the student answers, the LLM evaluates that answer (perhaps using a hidden rubric or via another query to a checking function) and then decides what to do next: praise and move on if correct (“Great, now what about the next part...”), or correct/misconception if wrong (“I see. There’s a small mistake in your approach regarding [...]; think about ... and try again.”), or give a hint if the student is silent or clearly stuck (“Maybe consider the formula for [...]; does that apply here?”). This adaptivity can be achieved by a combination of prompt engineering and possibly external logic (since current LLMs can be guided by a designed conversation flow using role instructions). Essentially, the LLM script becomes a structured tutor: identify current status -> respond with either hint, feedback, or next question. This realizes something like an Intelligent Tutoring System behavior but using the flexible language of an LLM. Importantly, scaffolding prompts should also fade assistance: as the student demonstrates competence, the LLM should step back. For example, after a few problems done with heavy guidance, the LLM might say, “Now, try the next one on your own; I’ll be here if you need help.” This encourages independence over time, addressing the dependency issue.

\subsection{Reflective Dialogue and Meta-Cognitive Prompts}
To promote deeper learning and self-regulation, the LLM can be prompted to incorporate reflection questions and prompts for elaboration. For example, after explaining a concept, the AI might ask the student, “Can you summarize this in your own words?” or “How would you apply this idea to a real-world scenario?” If the conversation is ending, the LLM could ask, “What was the most confusing part of this for you?” and then address that confusion. Another tactic: periodically have the LLM encourage the student to predict an answer before it’s given, e.g., “What do you think will happen if we do X?” before revealing the outcome. These strategies force the student to engage actively and self-reflect, which aligns with dialogic and constructivist practices. We can program these by simply adding to the LLM’s instructions: e.g., after every solution, ask the user a relevant follow-up question that checks understanding or connects to prior knowledge. There is evidence that self-explanation is a powerful learning technique; an AI that routinely asks for it leverages that effect.

\subsection{Persona and Tone Adjustments for Motivation}
One advantage of LLMs is the ability to adopt different tones or personas. While they lack true emotion, they can simulate encouragement or enthusiasm. A strategy could be to give the AI a motivational mentor persona – always patient, using positive reinforcement for effort (“I see you tried hard on that problem!”) and normalizing mistakes (“That’s a common misunderstanding, don’t worry, we can work through it.”). This can be done via a system prompt that sets the style: “You are a friendly tutor who always encourages the student and acknowledges their feelings.” Additionally, the AI can occasionally inject brief motivational phrases (“This is tough, but you’re making progress!”) or relate material to the student’s interests if known, to keep engagement high. Emotional support bots like in Cao et al. (2023) illustrate that explicitly separating this function is useful – in practice, an educational LLM could either integrate supportive feedback in its answers or have a dedicated mode the student can invoke when frustrated (even something like, “I need encouragement,” upon which the AI switches to a pep-talk mode). While this doesn’t give the AI true human empathy, it can at least provide some of the affective support that theories say is important for perseverance.

\subsection{Retrieval-Augmented Generation (RAG) for Factuality and Context}
Integrating external knowledge sources into LLM responses (RAG) is crucial for accuracy and alignment with curriculum. For educational uses, this could mean connecting the LLM to a database of vetted academic content – textbooks, lecture notes, or reference materials. When a student asks a question, the system can fetch relevant snippets (e.g., the relevant chapter section or a definition from a reliable source) and provide those as grounding for the LLM’s answer. This ensures that the conversation stays tethered to correct information and even specific class content (if the retrieval is from the course materials). RAG can also be used to reinforce learning strategies: for example, if a student asks a conceptual question, the system might retrieve a paragraph from the course text and prompt the LLM to ask the student to read or interpret that paragraph, rather than just hearing it from the AI. This engages the student with primary materials (promoting skills like reading comprehension and linking to authoritative sources). Technically, implementing RAG in an educational context would involve indexing all relevant course content or knowledge bases and using the student’s query (and possibly recent context of the conversation) as a search query.   The retrieved text can then be woven into the AI’s reply or provided as a quote with a citation. This method not only improves accuracy (reducing hallucinations) but also allows the AI to point students to additional resources (“You can find more on this in [textbook section], let’s read that excerpt together…”). It aligns with Laurillard’s idea of linking the student’s conceptual understanding to the formal knowledge represented in learning materials. Additionally, by citing sources, the AI can model good scholarly practice and encourage the student to trust-but-verify by checking original materials.  This is, in practice, not hard to do: the author has created tailored bespoke LLMs for his children who are currently taking final assessments before finishing to-16 education.  Each has the syllabus for the topic, past exam papers and answers for the past 5 years,  key texts, study guides and notes retrieved from the internet or school resource packs.  The LLM acts as a supportive tutor asking them questions and exploring the topic, and giving them feedback on structure and content for exam questions as it goes.  Each took less than 15 minutes to create.

Whilst searching for accuracy in the AI tutor is important, and reducing hallucinations a key element in this \citep{huang_survey_2025}, verifying correctness is also a fundamental knowledge skill that needs to be taught - even more so in these days of social media manipulation, deepfake technology, and diffusion-based image creation software.  Teaching students to not necessarily rely on all information given by an AI system but to check (especially references), or verify some medical advice with other sources, or at least engage the LLM in conversation to get it to justify and indicate its sources, is a necessary skill that all of use should learn.  Whilst we should make  attempts to engineer out such falsehoods in the AI through prompt engineering and RAG grounding in reliable sources, we must also encourage intellectual scepticism as a necessary feature of digital life.

Some falsehoods are helpful in education, however.  We also need to consider that LLMs will have to use different assumptions at different levels: for junior or high school science students, Newton's Laws are appropriate, but for University physicists, those truths become suddenly questioned and, at a basic level, fundamentally wrong.  But appropriate explanations to younger learners would retain them as facts, revising that view later as their knowledge developed.  This level of factual appropriateness is something that has to be given to the LLM in terms of the RAG or specifically scripted prompting.

\subsection{Structured Dialogue Flows (Hybrid AI-Human Design)}
Given the subtle unpredictability of LLMs, one strategy suggested  is to use a hybrid approach: have a rule-based or script-based structure for the conversation logic, but fill in the content with LLM generation where needed. For instance, a lesson could be structured as: Step 1: AI asks student 3 prerequisite questions (scripted questions). Step 2: Based on answers, AI (via rules) classifies understanding (good, medium, poor). Step 3: If poor, AI uses an LLM prompt to give a targeted explanation or analogy; if medium, AI asks an LLM to generate a challenging follow-up question; if good, AI proceeds to next topic. Step 4: AI presents a practice problem (perhaps retrieved from a database or generated and verified via LLM). Step 5: Student responds; AI (LLM) evaluates the response and gives feedback. And so on. In this setup, the pedagogical sequence and decisions (the structure) are controlled by a deterministic or semi-deterministic framework (which can be informed by educational best practices), while the flexibility and language richness of the LLM is used to execute the steps (posing questions, evaluating answers, giving hints). This mitigates the risk of the AI drifting off course or forgetting to do an important pedagogical step. Essentially, the AI is constrained to a conversational script that embodies a pedagogic strategy, but it fills that script with dynamic, personalized content. Tools or platforms could allow educators to design such flows without coding by specifying, e.g., “After teaching concept X, if student seems confused, provide a simpler example; if student seems confident, pose an extension question.” The LLM would then generate the example or question as needed. This addresses the limitation of LLMs not sticking to structure, noted in [8], by programmatically ensuring structure. It leverages strengths of both rule-based CAI (pedagogical control) and generative AI (personalization and natural interaction).

\subsection{Unstructured Dialogue}
We must recognise one of the unique capabilities of LLMs in any discussion on pedagogy - as well as being able to be supportive experts in the subject at hand, they also retain the abilities to converse knowledgeably on a wide range of subjects.  This flexibility can alleviate one of the issues identified above, namely the lack of dialogic depth.  By facilitating off-topic conversations and allowing in-depth explorations of tangential topics, the AI can support enquiry-based learning at a moment's notice.  If this is appropriately balanced so that conversations can be brought back to a specific topic under consideration over time, this ability to suddenly be side-tracked for a short while reflects many human-human conversations and represent a more holistic view of education and learning.  By supporting multiple threads of a conversation, the LLM encapsulates the different interests, projects, ideas and tasks people are typically juggling at any one time, and so becomes a more integrated partner in discussions around learning and life.

\subsection{Multi-Agent Simulations and Role-Plays}
Another novel strategy is to utilize LLMs’ ability to simulate roles to create richer learning scenarios. For example, for a debate or dialectical inquiry, one could have the LLM present two personas (perhaps even two separate LLM instances): one advocating position A, another advocating position B, and have the student moderate or participate. For instance, in an ethics class, the student could engage in a three-way conversation: the AI playing a “pro” stance, another AI a “con” stance, and the student weighs in or asks questions to each. This kind of role-play can expose students to multiple perspectives and teach argumentation. It’s similar to techniques used in classroom discussions, but here an AI can generate the contrasting viewpoints if peers aren’t available. Another scenario: historical role-play – the LLM is instructed to act as a historical figure answering questions in character (with factual grounding via RAG ideally). The student then “interviews”, say, Albert Einstein or Marie Curie. This makes the dialogue engaging and memorable. From a pedagogic standpoint, role-playing is linked to deeper engagement and empathy; an AI that can convincingly role-play can bring that into individual study. When doing this, guardrails are needed to keep the role factual (so the AI doesn’t distort the figure’s views), likely via RAG or curated knowledge about the figure.

\subsection{Periodic Knowledge Checks and Retrieval Practice}
To align with evidence-based learning strategies, LLM interactions can be interwoven with quizzes or recall prompts. For example, after a topic is taught in conversation, the AI could say, “Let’s do a quick check: [asks a question from earlier content].” If the student gets it right, great – practice successful. If not, the AI can revisit or clarify. This is essentially using the testing effect – the act of retrieving information helps solidify memory. Because an LLM can generate endless questions, it can tailor these to what was covered. Even simply asking “Earlier we discussed concept Y – can you recall the definition?” would prompt the student to retrieve it from memory. Such retrieval practice is typically planned by teachers; with an LLM, it can happen organically in the flow of conversation. A strategy could be: maintain a short memory of key points covered and randomly (or strategically) ask about them later in the chat. This also addresses that AI tends to always move forward; by intentionally looping back, we reinforce prior knowledge and ensure cumulative learning rather than one-off Q\&As.

\subsection{Transparent Explanations and Justifications}
Educators often highlight the importance of not just giving an answer but explaining the reasoning. LLMs often do produce explanations, but we can reinforce this by prompting the model to always justify its answers or steps. For instance, a template for the AI’s answers could be: First, restate the question in your own words, then break down the reasoning step by step, then arrive at the answer. This is similar to the “think-aloud” or “worked example” approach that benefits learners. Moreover, showing reasoning allows the student (and teacher) to spot if the AI’s reasoning has flaws. If an AI just says the answer without reasoning, a student might not trust or follow it; with reasoning, it models how one might think through the problem. This practice aligns with cognitive apprenticeship models where the expert demonstrates the thought process explicitly. An example: Student: “How does photosynthesis work?” AI (with template): “Okay, the question is asking how photosynthesis works – that is, how plants convert sunlight into energy. Let’s break it into steps. First, plants take in sunlight using chlorophyll... [continues through steps] ... finally, they produce glucose for energy. Therefore, photosynthesis works by...”. This thorough approach might be a bit verbose, but it ensures completeness. One might allow the student to interrupt or say “I know that part, you can skip ahead,” as a control for verbosity.  Some of the more recent models have a 'reasoning' mode in which their internal breakdown of the problem into steps much like these can be inspected, without being explicitly coded for in the prompt engineering.

\subsection{Personalization and Student Modelling}
To better hit the student’s ZPD, future systems might integrate a student model (either through explicit assessments or through analysis of past interactions). A simple method is ask the student about their background at the start. For instance, an AI tutor session could begin with the AI asking, “Before we begin, can you tell me what you already know about this topic or what level you’re at (beginner, intermediate)?”, or for the student to take a level assessement quiz beforehand.  This information can then be used to adjust the explanation detail. The AI could have different modes: e.g., if user says “I’m a high school student struggling with algebra,” the AI can use simpler language and more basic examples; if the user is a grad student wanting quick info, the AI might cut to the chase with more jargon. Prompting for this self-assessment not only helps tailor the interaction but also engages the learner in reflecting on their own state (which is itself a meta-cognitive act). Similarly, the AI could set up learning goals: “What would you like to achieve in this session? Solve a specific problem, or understand a concept, or something else?” By doing this, it aligns the conversation to the learner’s objectives. Such customization makes the dialogue more effective and respects the diversity of learners, something theories always emphasize.

Implementing these strategies can be done at different levels – some can be achieved immediately by prompt engineering and careful conversation design, others might require more complex system development (like a full RAG pipeline or multi-agent coordination). Nonetheless, even simple guidelines can be given to end-users (teachers or students) on how to prompt the AI for better educational outcomes. For example, teachers could share prompt templates with students like: “When you study with ChatGPT, try asking it to quiz you or to not give you the answer immediately but to give you hints.” Likewise, tool builders can integrate these as defaults (like a “tutor mode” in the interface that automatically does Socratic questioning and so on). The overall goal of these strategies is to ensure that the LLM acts in service of learning, rather than just information dispensing. They infuse the dialogue with elements of proven pedagogical techniques: questioning, scaffolding, feedback, practice, contextualization, and encouragement. By doing so, we approximate the behaviors of an ideal human tutor who is knowledgeable, patient, and pedagogically savvy – but with the added benefits of an AI’s breadth of knowledge and tireless availability. These adaptations also help to mitigate issues identified earlier: for instance, Socratic and scaffolded prompting counteracts the over-directness problem; RAG and justification address accuracy; motivational personas and reflection address the lack of human touch and meta-cognition, and structured flows keep the AI on a pedagogical track.

\subsection{The challenge of domains}
There are distinct differences in learning and teaching in different domains: in the sciences and maths, there are a lot of things that are either right or wrong, whereas this distinction is much harder to make in the humanities, where things are much less certain.  LLMs are inherently flexible and useful across these domains, and if constructed properly, can be effective in both - they can be oracles and checks for correctness in areas of precision and certainty; they can pose questions and provoke thought about alternative perspectives to encourage students to see different perspectives on a social or historical issue, for example.  This doesn't appear automatically: it is a characteristic of the pedagogical strategies we guide the LLMs towards, and different strategies will be better suited to some domains than others.  Some of the exciting possibilities for LLMs in education are away from the precision of coding something or solving a maths equation (areas in which they tend to have been applied more) and into the humanities and arts.  They can ignite curiosity by asking penetrating questions, encouraging students to explore ideas around events in history, politics or literature in ways that other tutoring systems haven't had the flexibility and generality to cover before.

\section{An AI Ecosystem}

One of the consequences of the ubiquitous nature of LLMs is that any new pedagogical strategies we try to impose on the AI assistants are sitting with a more complex ecosystem than previously envisaged. When coding, for example, students may be using CoPilot within their programming environment to help write code, whilst simultaneously asking for details on API calls or data structures from a generic ChatGPT model.   This provides more support for the general argument in this paper, that a conversational, ongoing, dialogic approach to interaction is required.  These new forms of model support serendipitous explorations of topics of interest, interwoven with focussed information seeking activities, general knowledge-building materials, or quizzes and questioning to cement knowledge in long-term memory.  We will develop a relationship with the AI systems across many facets of our lives that extend far beyond anything ever imagined when people created the early intelligent tutoring systems, and our expectations of educational systems will be shaped by these interactions too.  We are likely to require our AI assistants to be flexible - to be able to ask a programming tutor for help with a Physics problem, or a relationship issue, is not going to be out of the ordinary. This generality of purpose can be handled in multiple ways - splitting different functionalities into different roles as per \citet{cao_ai_2023}, via multi-agent coordination approaches \citep{tran_multi-agent_2025}, or thorough sociotechnical means such as having trusted models to turn to for specific tasks - much as academic scholars look to Google Scholar for research articles, so for example programmers may turn to Claude Sonnet 3.7 for code completion and ChatGPT4.5 for general research.

One other consequence for learning, if LLMs become as widespread in usage as we predict, is that interaction as a whole will be redefined.  Instead of the common point-and-click interfaces, or scroll-and-press if on smartphones, the generality and flexibility of the conversational interactions will become the norm.  This is likely to lead to a demand for much more conversational approaches across all forms of digital life, and hence the structure of many existing systems and approaches will be challenged if they do not support this.  Thus it is not only pedagogical approaches that may have to change - whole models of digital education may need altering to fit the new expectations, even demands, of learners.

\section{Conclusion}

Conversational AI in the form of large language models has opened a new frontier in education. Dialogic pedagogy, once constrained to human-to-human interaction, can now be extended into human-AI dialogues that are capable of rich, contextual, and interactive communication. This research surveyed the emerging landscape of LLM applications in education, examined them through the lens of established learning theories, and discussed strategies to better align practice with pedagogy. The literature makes clear that LLMs and chatbots have demonstrated significant potential: they can act as always-available tutors, adapt to individual queries, and engage students in ways that foster understanding and motivation. Studies have reported positive learning outcomes and student receptiveness when these agents are used appropriately, especially as supplements to traditional teaching.  We are potentially close to realising the benefits of Bloom's 2 sigma concept, where everyone can become a prodigy and experience highly engaging, highly personalised learning experiences to elevate their understanding far above previous expectations.  At the same time, we have seen that the integration of LLMs into learning is not without pitfalls. Traditional pedagogical frameworks did not account for an omnipresent, knowledge-generating entity, and so we find mismatches – from the risk of shortcutting student thinking to concerns about factual accuracy and the nuances of human touch. By mapping theories like Socratic dialogue, scaffolding, and conversational frameworks onto what LLMs can (and cannot) do, we identified where the gaps lie. Notably, the importance of maintaining a learner’s active role and agency emerged repeatedly: without intentional design, an AI might encourage passivity or dependence, which educational theory warns against. Encouragingly, many of these challenges can be addressed through deliberate pedagogical design of AI interactions. We outlined a set of concrete strategies – essentially a toolkit for educators and developers – to guide LLMs toward being better tutors. These ranged from micro-level tactics (like how to phrase prompts to elicit a Socratic style) to system-level architectures (like hybrid scripted-AI systems and retrieval augmentation). What these interventions have in common is that they put pedagogical intent front and center, using the AI’s flexibility to implement known good practices. Early experiments, such as Cao et al.’s multi-role chatbot environment, show that when theory-informed design is applied, students respond with greater engagement and motivation, validating the approach. For higher education, where learners often pursue knowledge independently and critically, LLMs – configured properly – can serve as intellectual partners that challenge and support learners akin to a mentor or Socratic professor. We have also noted implications for secondary education: structured AI tutors could help fill gaps in resources, providing individualized attention that some school systems lack, though care must be taken to ensure alignment with curricular goals and developmental appropriateness. In lifelong learning, adult learners stand to benefit from AI assistants that adapt to their goals, whether it’s upskilling for a job or exploring a new hobby, especially if those AIs are tuned to respect adult learning principles (autonomy, relevancy, drawing on experience). It is important to emphasize that LLM-based education is not a replacement for human educators or peers, but a complement. The best outcomes likely arise when human and AI strengths are combined – for example, an instructor devises the learning plan and provides emotional intelligence and ethical oversight, while the LLM offers on-demand tutoring and practice opportunities. Human teachers can also use LLMs as a tool (for generating examples, providing feedback on drafts, etc.), thus extending their own capabilities. This collaborative model aligns with Sætra’s idea that the most powerful scaffold is AI plus human peers/instructors together.  

Our conclusion is that a conversational approach to pedagogy is vital to maximise the capabilities of LLMs, recognising that learning continues over a longer period of time in an interactive manner, across both formal and informal settings.  Coupled with appropriate prompting and RAG, this strategy leverages the flexibility and generality of LLMs whilst anchoring them in sound factual bases and a deep awareness of Socratic, reflective dialogue.

Future research emerging from our work and the others discussed here include conducting controlled studies on the effectiveness of the proposed prompting and RAG strategies on learning outcomes; exploring ways to better model and detect a student’s state so that AI responses can be more finely tuned; developing domain-specific LLM fine-tunes that encapsulate pedagogical content knowledge (so that the model not only knows the subject, but common misconceptions and how to address them); and investigating the long-term impact of learning with AI tutors on skills like self-regulation, critical thinking, and creativity. There is also a need for frameworks to evaluate AI tutors not just on accuracy, but on pedagogical criteria – for example, metrics for “did the AI keep the student engaged in higher-order thinking?” or “did the AI’s hints lead the student to an aha-moment or just give away the answer?”. In conclusion, LLMs are catalyzing a renaissance of conversational learning, compelling educators to revisit and adapt dialogic pedagogy for the digital age. If we intentionally infuse our interactions with these models with the wisdom of educational theories – turning them into Socratic guides, patient scaffolders, and reflective confidants – we stand to unlock truly transformative learning experiences. The dialogue between student and AI can then become not a gimmick or a shortcut, but a genuine space of learning: a space where questions are valued as much as answers, where mistakes are stepping stones, and where knowledge is co-constructed in the interplay of human curiosity and machine intelligence. With thoughtful integration, LLMs can help realize the long-held dream of education that is at once deeply personalized and richly dialogic, empowering learners of all ages to reach their potential in conversation with the collective knowledge of humanity.

\section*{Acknowledgements}
I'd like to thank Anon for her insightful comments. Generative AI made the following contributions: the author has used them to enhance his coding (Claude Sonnet 3.5/3.7, CoPilot, ChatGPT o3 -mini-high), to learn new skills (ChatGPT3.0/4.0/4.5, Gemini), and he has created bespoke models with them to experience how they teach and retain context and awareness (ChatGPT custom).  Some of that experiential learning was undertaking tasks in the preparation of this manuscript, ranging from identifying possible sources to converting references from plain text into BiB\TeX.  Any text produced has been re-written and edited by the author into his own style and he bears responsibility for it.

\section*{Declarations}

\begin{itemize}
\item Funding: none to declare
\item Conflict of interest: none
\item Ethics approval and consent to participate: all relevant ethical clearances were obtained.
\item Consent for publication: no consensts required
\item Data availability: none generated
\item Materials availability: none generated
\item Code availability: none generated
\item Author contribution: RB conceived, developed and produced the manuscript
\end{itemize}

\bibliography{references}

\begin{thebibliography}{}
\renewcommand{\doi}[1]{\url{https://doi.org/#1}}
\bibcommenthead

\bibitem [\protect \citeauthoryear {%
Alexander%
}{%
Alexander%
}{%
{\protect \APACyear {2008}}%
}]{%
alexander2008culture}
\APACinsertmetastar {%
alexander2008culture}%
\begin{APACrefauthors}%
Alexander, R.%
\end{APACrefauthors}%
\unskip\
\newblock
\APACrefYearMonthDay{2008}{}{}.
\newblock
{\BBOQ}\APACrefatitle {Culture, dialogue and learning: {Notes} on an emerging pedagogy} {Culture, dialogue and learning: {Notes} on an emerging pedagogy}.{\BBCQ}
\newblock
\APACjournalVolNumPages{Exploring talk in school}{2008}{}{91--114,}
\newblock

\newblock

\PrintBackRefs{\CurrentBib}

\bibitem [\protect \citeauthoryear {%
Becker%
\ \protect \BOthers {.}}{%
Becker%
\ \protect \BOthers {.}}{%
{\protect \APACyear {2023}}%
}]{%
becker_programming_2023}
\APACinsertmetastar {%
becker_programming_2023}%
\begin{APACrefauthors}%
Becker, B.A.%
, Denny, P.%
, Finnie-Ansley, J.%
, Luxton-Reilly, A.%
, Prather, J.%
\BCBL {} Santos, E.A.%
\end{APACrefauthors}%
\unskip\
\newblock
\APACrefYearMonthDay{2023}{{\APACmonth{03}}}{}.
\newblock
{\BBOQ}\APACrefatitle {Programming {Is} {Hard} - {Or} at {Least} {It} {Used} to {Be}: {Educational} {Opportunities} and {Challenges} of {AI} {Code} {Generation}} {Programming {Is} {Hard} - {Or} at {Least} {It} {Used} to {Be}: {Educational} {Opportunities} and {Challenges} of {AI} {Code} {Generation}}.{\BBCQ}
\newblock
 \APACrefbtitle {Proceedings of the 54th {ACM} {Technical} {Symposium} on {Computer} {Science} {Education} {V}. 1} {Proceedings of the 54th {ACM} {Technical} {Symposium} on {Computer} {Science} {Education} {V}. 1}\ (\BPGS\ 500--506).
\newblock
\APACaddressPublisher{Toronto ON Canada}{ACM}.
\newblock
\begin{APACrefURL} [{2025-04-04}]{https://dl.acm.org/doi/10.1145/3545945.3569759} \end{APACrefURL}
\PrintBackRefs{\CurrentBib}

\bibitem [\protect \citeauthoryear {%
Bird%
\ \protect \BOthers {.}}{%
Bird%
\ \protect \BOthers {.}}{%
{\protect \APACyear {2023}}%
}]{%
bird_taking_2023}
\APACinsertmetastar {%
bird_taking_2023}%
\begin{APACrefauthors}%
Bird, C.%
, Ford, D.%
, Zimmermann, T.%
, Forsgren, N.%
, Kalliamvakou, E.%
, Lowdermilk, T.%
\BCBL {} Gazit, I.%
\end{APACrefauthors}%
\unskip\
\newblock
\APACrefYearMonthDay{2023}{{\APACmonth{01}}}{}.
\newblock
{\BBOQ}\APACrefatitle {Taking {Flight} with {Copilot}: {Early} insights and opportunities of {AI}-powered pair-programming tools} {Taking {Flight} with {Copilot}: {Early} insights and opportunities of {AI}-powered pair-programming tools}.{\BBCQ}
\newblock
\APACjournalVolNumPages{Queue}{20}{6}{Pages 10:35--Pages 10:57,}
\newblock
\begin{APACrefDOI} \doi{10.1145/3582083} \end{APACrefDOI}
\newblock
\begin{APACrefURL} [{2025-04-04}]{https://dl.acm.org/doi/10.1145/3582083} \end{APACrefURL}
\newblock

\newblock

\PrintBackRefs{\CurrentBib}

\bibitem [\protect \citeauthoryear {%
Bloom%
}{%
Bloom%
}{%
{\protect \APACyear {1984}}%
}]{%
bloom_2_1984}
\APACinsertmetastar {%
bloom_2_1984}%
\begin{APACrefauthors}%
Bloom, B.S.%
\end{APACrefauthors}%
\unskip\
\newblock
\APACrefYearMonthDay{1984}{}{}.
\newblock
{\BBOQ}\APACrefatitle {The 2 sigma problem: {The} search for methods of group instruction as effective as one-to-one tutoring} {The 2 sigma problem: {The} search for methods of group instruction as effective as one-to-one tutoring}.{\BBCQ}
\newblock
\APACjournalVolNumPages{Educational researcher}{13}{6}{4--16,}
\newblock
\APACrefnote{Publisher: Sage Publications Sage CA: Thousand Oaks, CA}
\newblock

\newblock

\PrintBackRefs{\CurrentBib}

\bibitem [\protect \citeauthoryear {%
Blythe%
}{%
Blythe%
}{%
{\protect \APACyear {2023}}%
}]{%
blythe_artificial_2023}
\APACinsertmetastar {%
blythe_artificial_2023}%
\begin{APACrefauthors}%
Blythe, M.%
\end{APACrefauthors}%
\unskip\
\newblock
\APACrefYearMonthDay{2023}{{\APACmonth{11}}}{}.
\newblock
{\BBOQ}\APACrefatitle {Artificial {Design} {Fiction}: {Using} {AI} as a {Material} for {Pastiche} {Scenarios}} {Artificial {Design} {Fiction}: {Using} {AI} as a {Material} for {Pastiche} {Scenarios}}.{\BBCQ}
\newblock
 \APACrefbtitle {Proceedings of the 26th {International} {Academic} {Mindtrek} {Conference}} {Proceedings of the 26th {International} {Academic} {Mindtrek} {Conference}}\ (\BPGS\ 195--206).
\newblock
\APACaddressPublisher{New York, NY, USA}{Association for Computing Machinery}.
\newblock
\begin{APACrefURL} [{2025-04-14}]{https://doi.org/10.1145/3616961.3616987} \end{APACrefURL}
\PrintBackRefs{\CurrentBib}

\bibitem [\protect \citeauthoryear {%
Bui%
, Collier%
, Ozturk%
\BCBL {}\ \BBA {} Song%
}{%
Bui%
\ \protect \BOthers {.}}{%
{\protect \APACyear {2025}}%
}]{%
bui_effects_2025}
\APACinsertmetastar {%
bui_effects_2025}%
\begin{APACrefauthors}%
Bui, N.%
, Collier, J.%
, Ozturk, Y.E.%
\BCBL {} Song, D.%
\end{APACrefauthors}%
\unskip\
\newblock
\APACrefYearMonthDay{2025}{{\APACmonth{04}}}{}.
\newblock
{\BBOQ}\APACrefatitle {The {Effects} of {Conversational} {Agents} on {Human} {Learning} and {How} {We} {Used} {Them}: {A} {Systematic} {Review} of {Studies} {Conducted} {Before} {Generative} {AI}} {The {Effects} of {Conversational} {Agents} on {Human} {Learning} and {How} {We} {Used} {Them}: {A} {Systematic} {Review} of {Studies} {Conducted} {Before} {Generative} {AI}}.{\BBCQ}
\newblock
\APACjournalVolNumPages{TechTrends}{}{}{,}
\newblock
\begin{APACrefDOI} \doi{10.1007/s11528-025-01066-0} \end{APACrefDOI}
\newblock
\begin{APACrefURL} [{2025-04-09}]{https://doi.org/10.1007/s11528-025-01066-0} \end{APACrefURL}
\newblock

\newblock

\PrintBackRefs{\CurrentBib}

\bibitem [\protect \citeauthoryear {%
Cao%
, Ding%
, Lin%
\BCBL {}\ \BBA {} Hopfgartner%
}{%
Cao%
\ \protect \BOthers {.}}{%
{\protect \APACyear {2023}}%
}]{%
cao_ai_2023}
\APACinsertmetastar {%
cao_ai_2023}%
\begin{APACrefauthors}%
Cao, C.C.%
, Ding, Z.%
, Lin, J.%
\BCBL {} Hopfgartner, F.%
\end{APACrefauthors}%
\unskip\
\newblock
\APACrefYearMonthDay{2023}{{\APACmonth{08}}}{}.
\newblock
\APACrefbtitle {{AI} {Chatbots} as {Multi}-{Role} {Pedagogical} {Agents}: {Transforming} {Engagement} in {CS} {Education}.} {{AI} {Chatbots} as {Multi}-{Role} {Pedagogical} {Agents}: {Transforming} {Engagement} in {CS} {Education}.}
\newblock
\APACaddressPublisher{}{arXiv}.
\newblock
\begin{APACrefURL} [{2025-04-09}]{http://arxiv.org/abs/2308.03992} \end{APACrefURL}
\newblock
\APACrefnote{arXiv:2308.03992 [cs]}
\PrintBackRefs{\CurrentBib}

\bibitem [\protect \citeauthoryear {%
Chang%
}{%
Chang%
}{%
{\protect \APACyear {2023}}%
}]{%
chang2023prompting}
\APACinsertmetastar {%
chang2023prompting}%
\begin{APACrefauthors}%
Chang, E.Y.%
\end{APACrefauthors}%
\unskip\
\newblock
\APACrefYearMonthDay{2023}{}{}.
\newblock
{\BBOQ}\APACrefatitle {Prompting large language models with the socratic method} {Prompting large language models with the socratic method}.{\BBCQ}
\newblock
 \APACrefbtitle {{IEEE} {CCWC} 2023 (proceedings of the computing and communication workshop and conference).} {{IEEE} {CCWC} 2023 (proceedings of the computing and communication workshop and conference).}
\PrintBackRefs{\CurrentBib}

\bibitem [\protect \citeauthoryear {%
{Chris Rowlands}%
}{%
{Chris Rowlands}%
}{%
{\protect \APACyear {2025}}%
}]{%
chris_rowlands_goodbye_2025}
\APACinsertmetastar {%
chris_rowlands_goodbye_2025}%
\begin{APACrefauthors}%
{Chris Rowlands}%
\end{APACrefauthors}%
\unskip\
\newblock
\APACrefYearMonthDay{2025}{{\APACmonth{02}}}{}.
\newblock
\APACrefbtitle {Goodbye {Google}? {People} are increasingly switching to the likes of {ChatGPT}, according to major survey – here’s why.} {Goodbye {Google}? {People} are increasingly switching to the likes of {ChatGPT}, according to major survey – here’s why.}
\newblock
\begin{APACrefURL} [{2025-04-14}]{https://www.techradar.com/tech/people-are-increasingly-swapping-google-for-the-likes-of-chatgpt-according-to-a-major-survey-heres-why} \end{APACrefURL}
\PrintBackRefs{\CurrentBib}

\bibitem [\protect \citeauthoryear {%
Freire%
}{%
Freire%
}{%
{\protect \APACyear {2000}}%
}]{%
freire_pedagogy_2000}
\APACinsertmetastar {%
freire_pedagogy_2000}%
\begin{APACrefauthors}%
Freire, P.%
\end{APACrefauthors}%
\unskip\
\newblock
\APACrefYear{2000}.
\newblock
\APACrefbtitle {Pedagogy of the oppressed} {Pedagogy of the oppressed}\ (\PrintOrdinal{30th anniversary ed}\ \BEd).
\newblock
\APACaddressPublisher{New York}{Continuum}.
\PrintBackRefs{\CurrentBib}

\bibitem [\protect \citeauthoryear {%
Han%
, Lee%
\BCBL {}\ \BBA {} Lee%
}{%
Han%
\ \protect \BOthers {.}}{%
{\protect \APACyear {2010}}%
}]{%
han_impact_2010}
\APACinsertmetastar {%
han_impact_2010}%
\begin{APACrefauthors}%
Han, K\BHBI W.%
, Lee, E.%
\BCBL {} Lee, Y.%
\end{APACrefauthors}%
\unskip\
\newblock
\APACrefYearMonthDay{2010}{{\APACmonth{05}}}{}.
\newblock
{\BBOQ}\APACrefatitle {The {Impact} of a {Peer}-{Learning} {Agent} {Based} on {Pair} {Programming} in a {Programming} {Course}} {The {Impact} of a {Peer}-{Learning} {Agent} {Based} on {Pair} {Programming} in a {Programming} {Course}}.{\BBCQ}
\newblock
\APACjournalVolNumPages{IEEE Transactions on Education}{53}{2}{318--327,}
\newblock
\begin{APACrefDOI} \doi{10.1109/TE.2009.2019121} \end{APACrefDOI}
\newblock
\begin{APACrefURL} [{2025-04-04}]{https://ieeexplore.ieee.org/abstract/document/5196685} \end{APACrefURL}
\newblock
\APACrefnote{Conference Name: IEEE Transactions on Education}
\newblock

\newblock

\PrintBackRefs{\CurrentBib}

\bibitem [\protect \citeauthoryear {%
Huang%
\ \protect \BOthers {.}}{%
Huang%
\ \protect \BOthers {.}}{%
{\protect \APACyear {2025}}%
}]{%
huang_survey_2025}
\APACinsertmetastar {%
huang_survey_2025}%
\begin{APACrefauthors}%
Huang, L.%
, Yu, W.%
, Ma, W.%
, Zhong, W.%
, Feng, Z.%
, Wang, H.%
\BDBL {}Liu, T.%
\end{APACrefauthors}%
\unskip\
\newblock
\APACrefYearMonthDay{2025}{{\APACmonth{01}}}{}.
\newblock
{\BBOQ}\APACrefatitle {A {Survey} on {Hallucination} in {Large} {Language} {Models}: {Principles}, {Taxonomy}, {Challenges}, and {Open} {Questions}} {A {Survey} on {Hallucination} in {Large} {Language} {Models}: {Principles}, {Taxonomy}, {Challenges}, and {Open} {Questions}}.{\BBCQ}
\newblock
\APACjournalVolNumPages{ACM Trans. Inf. Syst.}{43}{2}{42:1--42:55,}
\newblock
\begin{APACrefDOI} \doi{10.1145/3703155} \end{APACrefDOI}
\newblock
\begin{APACrefURL} [{2025-04-01}]{https://doi.org/10.1145/3703155} \end{APACrefURL}
\newblock

\newblock

\PrintBackRefs{\CurrentBib}

\bibitem [\protect \citeauthoryear {%
Jarratt%
, Bowman%
, Culver%
\BCBL {}\ \BBA {} Segre%
}{%
Jarratt%
\ \protect \BOthers {.}}{%
{\protect \APACyear {2019}}%
}]{%
jarratt_large-scale_2019}
\APACinsertmetastar {%
jarratt_large-scale_2019}%
\begin{APACrefauthors}%
Jarratt, L.%
, Bowman, N.A.%
, Culver, K.%
\BCBL {} Segre, A.M.%
\end{APACrefauthors}%
\unskip\
\newblock
\APACrefYearMonthDay{2019}{{\APACmonth{07}}}{}.
\newblock
{\BBOQ}\APACrefatitle {A {Large}-{Scale} {Experimental} {Study} of {Gender} and {Pair} {Composition} in {Pair} {Programming}} {A {Large}-{Scale} {Experimental} {Study} of {Gender} and {Pair} {Composition} in {Pair} {Programming}}.{\BBCQ}
\newblock
 \APACrefbtitle {Proceedings of the 2019 {ACM} {Conference} on {Innovation} and {Technology} in {Computer} {Science} {Education}} {Proceedings of the 2019 {ACM} {Conference} on {Innovation} and {Technology} in {Computer} {Science} {Education}}\ (\BPGS\ 176--181).
\newblock
\APACaddressPublisher{New York, NY, USA}{Association for Computing Machinery}.
\newblock
\begin{APACrefURL} [{2025-04-14}]{https://dl.acm.org/doi/10.1145/3304221.3319782} \end{APACrefURL}
\PrintBackRefs{\CurrentBib}

\bibitem [\protect \citeauthoryear {%
Knowles%
}{%
Knowles%
}{%
{\protect \APACyear {1980}}%
}]{%
knowles_modern_1980}
\APACinsertmetastar {%
knowles_modern_1980}%
\begin{APACrefauthors}%
Knowles, M.S.%
\end{APACrefauthors}%
\unskip\
\newblock
\APACrefYear{1980}.
\newblock
\APACrefbtitle {The {Modern} {Practice} of {Adult} {Education}: {From} {Pedagogy} to {Andragogy}} {The {Modern} {Practice} of {Adult} {Education}: {From} {Pedagogy} to {Andragogy}}.
\newblock
\APACaddressPublisher{}{Association Press}.
\newblock
\APACrefnote{Google-Books-ID: \_gifAAAAMAAJ}
\PrintBackRefs{\CurrentBib}

\bibitem [\protect \citeauthoryear {%
Kuttal%
, Ong%
, Kwasny%
\BCBL {}\ \BBA {} Robe%
}{%
Kuttal%
\ \protect \BOthers {.}}{%
{\protect \APACyear {2021}}%
}]{%
kuttal_trade-offs_2021}
\APACinsertmetastar {%
kuttal_trade-offs_2021}%
\begin{APACrefauthors}%
Kuttal, S.K.%
, Ong, B.%
, Kwasny, K.%
\BCBL {} Robe, P.%
\end{APACrefauthors}%
\unskip\
\newblock
\APACrefYearMonthDay{2021}{{\APACmonth{05}}}{}.
\newblock
{\BBOQ}\APACrefatitle {Trade-offs for {Substituting} a {Human} with an {Agent} in a {Pair} {Programming} {Context}: {The} {Good}, the {Bad}, and the {Ugly}} {Trade-offs for {Substituting} a {Human} with an {Agent} in a {Pair} {Programming} {Context}: {The} {Good}, the {Bad}, and the {Ugly}}.{\BBCQ}
\newblock
 \APACrefbtitle {Proceedings of the 2021 {CHI} {Conference} on {Human} {Factors} in {Computing} {Systems}} {Proceedings of the 2021 {CHI} {Conference} on {Human} {Factors} in {Computing} {Systems}}\ (\BPGS\ 1--20).
\newblock
\APACaddressPublisher{New York, NY, USA}{Association for Computing Machinery}.
\newblock
\begin{APACrefURL} [{2025-04-04}]{https://dl.acm.org/doi/10.1145/3411764.3445659} \end{APACrefURL}
\PrintBackRefs{\CurrentBib}

\bibitem [\protect \citeauthoryear {%
Labadze%
, Grigolia%
\BCBL {}\ \BBA {} Machaidze%
}{%
Labadze%
\ \protect \BOthers {.}}{%
{\protect \APACyear {2023}}%
}]{%
labadze_role_2023}
\APACinsertmetastar {%
labadze_role_2023}%
\begin{APACrefauthors}%
Labadze, L.%
, Grigolia, M.%
\BCBL {} Machaidze, L.%
\end{APACrefauthors}%
\unskip\
\newblock
\APACrefYearMonthDay{2023}{}{}.
\newblock
{\BBOQ}\APACrefatitle {Role of {AI} {Chatbots} in {Education}: {A} {Systematic} {Literature} {Review}} {Role of {AI} {Chatbots} in {Education}: {A} {Systematic} {Literature} {Review}}.{\BBCQ}
\newblock
\APACjournalVolNumPages{International Journal of Educational Technology in Higher Education}{20}{56}{,}
\newblock
\begin{APACrefDOI} \doi{10.1186/s41239-023-00426-1} \end{APACrefDOI}
\newblock
\begin{APACrefURL} {https://doi.org/10.1186/s41239-023-00426-1} \end{APACrefURL}
\newblock

\newblock

\PrintBackRefs{\CurrentBib}

\bibitem [\protect \citeauthoryear {%
Laurillard%
}{%
Laurillard%
}{%
{\protect \APACyear {2013}}%
}]{%
laurillard_rethinking_2013}
\APACinsertmetastar {%
laurillard_rethinking_2013}%
\begin{APACrefauthors}%
Laurillard, D.%
\end{APACrefauthors}%
\unskip\
\newblock
\APACrefYear{2013}.
\newblock
\APACrefbtitle {Rethinking {University} {Teaching}: {A} {Conversational} {Framework} for the {Effective} {Use} of {Learning} {Technologies}} {Rethinking {University} {Teaching}: {A} {Conversational} {Framework} for the {Effective} {Use} of {Learning} {Technologies}}.
\newblock
\APACaddressPublisher{}{Routledge}.
\newblock
\APACrefnote{Google-Books-ID: H3VtZoM5vc8C}
\PrintBackRefs{\CurrentBib}

\bibitem [\protect \citeauthoryear {%
Lee%
\ \protect \BOthers {.}}{%
Lee%
\ \protect \BOthers {.}}{%
{\protect \APACyear {2024}}%
}]{%
lee_can_2024}
\APACinsertmetastar {%
lee_can_2024}%
\begin{APACrefauthors}%
Lee, U.%
, Jeong, Y.%
, Koh, J.%
, Byun, G.%
, Lee, Y.%
, Hwang, Y.%
\BDBL {}Lim, C.%
\end{APACrefauthors}%
\unskip\
\newblock
\APACrefYearMonthDay{2024}{}{}.
\newblock
{\BBOQ}\APACrefatitle {Can {ChatGPT} be a debate partner? {Developing} {ChatGPT}-based application “{DEBO}” for debate education, findings and limitations} {Can {ChatGPT} be a debate partner? {Developing} {ChatGPT}-based application “{DEBO}” for debate education, findings and limitations}.{\BBCQ}
\newblock
\APACjournalVolNumPages{Educational Technology \& Society}{27}{2}{pp. 321--346,}
\newblock
\begin{APACrefURL} [{2025-03-25}]{https://www.jstor.org/stable/48766178} \end{APACrefURL}
\newblock
\APACrefnote{Publisher: International Forum of Educational Technology \& Society, National Taiwan Normal University, Taiwan}
\newblock

\newblock

\PrintBackRefs{\CurrentBib}

\bibitem [\protect \citeauthoryear {%
Luccioni%
, Jernite%
\BCBL {}\ \BBA {} Strubell%
}{%
Luccioni%
\ \protect \BOthers {.}}{%
{\protect \APACyear {2024}}%
}]{%
luccioni_power_2024}
\APACinsertmetastar {%
luccioni_power_2024}%
\begin{APACrefauthors}%
Luccioni, S.%
, Jernite, Y.%
\BCBL {} Strubell, E.%
\end{APACrefauthors}%
\unskip\
\newblock
\APACrefYearMonthDay{2024}{{\APACmonth{06}}}{}.
\newblock
{\BBOQ}\APACrefatitle {Power {Hungry} {Processing}: {Watts} {Driving} the {Cost} of {AI} {Deployment}?} {Power {Hungry} {Processing}: {Watts} {Driving} the {Cost} of {AI} {Deployment}?}{\BBCQ}
\newblock
 \APACrefbtitle {Proceedings of the 2024 {ACM} {Conference} on {Fairness}, {Accountability}, and {Transparency}} {Proceedings of the 2024 {ACM} {Conference} on {Fairness}, {Accountability}, and {Transparency}}\ (\BPGS\ 85--99).
\newblock
\APACaddressPublisher{New York, NY, USA}{Association for Computing Machinery}.
\newblock
\begin{APACrefURL} [{2025-04-17}]{https://dl.acm.org/doi/10.1145/3630106.3658542} \end{APACrefURL}
\PrintBackRefs{\CurrentBib}

\bibitem [\protect \citeauthoryear {%
Ma%
, Shen%
, Koedinger%
\BCBL {}\ \BBA {} Wu%
}{%
Ma%
\ \protect \BOthers {.}}{%
{\protect \APACyear {2024}}%
}]{%
ma_how_2024}
\APACinsertmetastar {%
ma_how_2024}%
\begin{APACrefauthors}%
Ma, Q.%
, Shen, H.%
, Koedinger, K.%
\BCBL {} Wu, S.T.%
\end{APACrefauthors}%
\unskip\
\newblock
\APACrefYearMonthDay{2024}{}{}.
\newblock
{\BBOQ}\APACrefatitle {How to {Teach} {Programming} in the {AI} {Era}? {Using} {LLMs} as a {Teachable} {Agent} for {Debugging}} {How to {Teach} {Programming} in the {AI} {Era}? {Using} {LLMs} as a {Teachable} {Agent} for {Debugging}}.{\BBCQ}
\newblock
 A.M.~Olney, I\BHBI A.~Chounta, Z.~Liu, O.C.~Santos\BCBL {}\ \BBA {} I.I.~Bittencourt\ (\BEDS), \APACrefbtitle {Artificial {Intelligence} in {Education}} {Artificial {Intelligence} in {Education}}\ (\BPGS\ 265--279).
\newblock
\APACaddressPublisher{Cham}{Springer Nature Switzerland}.
\PrintBackRefs{\CurrentBib}

\bibitem [\protect \citeauthoryear {%
Manfredi%
, Erra%
\BCBL {}\ \BBA {} Gilio%
}{%
Manfredi%
\ \protect \BOthers {.}}{%
{\protect \APACyear {2023}}%
}]{%
manfredi_mixed_2023}
\APACinsertmetastar {%
manfredi_mixed_2023}%
\begin{APACrefauthors}%
Manfredi, G.%
, Erra, U.%
\BCBL {} Gilio, G.%
\end{APACrefauthors}%
\unskip\
\newblock
\APACrefYearMonthDay{2023}{{\APACmonth{06}}}{}.
\newblock
{\BBOQ}\APACrefatitle {A {Mixed} {Reality} {Approach} for {Innovative} {Pair} {Programming} {Education} with a {Conversational} {AI} {Virtual} {Avatar}} {A {Mixed} {Reality} {Approach} for {Innovative} {Pair} {Programming} {Education} with a {Conversational} {AI} {Virtual} {Avatar}}.{\BBCQ}
\newblock
 \APACrefbtitle {Proceedings of the 27th {International} {Conference} on {Evaluation} and {Assessment} in {Software} {Engineering}} {Proceedings of the 27th {International} {Conference} on {Evaluation} and {Assessment} in {Software} {Engineering}}\ (\BPGS\ 450--454).
\newblock
\APACaddressPublisher{New York, NY, USA}{Association for Computing Machinery}.
\newblock
\begin{APACrefURL} [{2025-04-04}]{https://dl.acm.org/doi/10.1145/3593434.3593952} \end{APACrefURL}
\PrintBackRefs{\CurrentBib}

\bibitem [\protect \citeauthoryear {%
Mercer%
}{%
Mercer%
}{%
{\protect \APACyear {2004}}%
}]{%
mercer_sociocultural_2004}
\APACinsertmetastar {%
mercer_sociocultural_2004}%
\begin{APACrefauthors}%
Mercer, N.%
\end{APACrefauthors}%
\unskip\
\newblock
\APACrefYearMonthDay{2004}{{\APACmonth{06}}}{}.
\newblock
{\BBOQ}\APACrefatitle {Sociocultural discourse analysis} {Sociocultural discourse analysis}.{\BBCQ}
\newblock
\APACjournalVolNumPages{Journal of Applied Linguistics}{1}{2}{137--168,}
\newblock
\begin{APACrefDOI} \doi{10.1558/japl.v1.i2.137} \end{APACrefDOI}
\newblock
\begin{APACrefURL} [{2025-04-17}]{https://utppublishing.com/doi/abs/10.1558/japl.v1.i2.137} \end{APACrefURL}
\newblock
\APACrefnote{Publisher: University of Toronto Press}
\newblock

\newblock

\PrintBackRefs{\CurrentBib}

\bibitem [\protect \citeauthoryear {%
Mercer%
\ \BBA {} Howe%
}{%
Mercer%
\ \BBA {} Howe%
}{%
{\protect \APACyear {2012}}%
}]{%
mercer_explaining_2012}
\APACinsertmetastar {%
mercer_explaining_2012}%
\begin{APACrefauthors}%
Mercer, N.%
\BCBT {}\ \BBA {} Howe, C.%
\end{APACrefauthors}%
\unskip\
\newblock
\APACrefYearMonthDay{2012}{{\APACmonth{03}}}{}.
\newblock
{\BBOQ}\APACrefatitle {Explaining the dialogic processes of teaching and learning: {The} value and potential of sociocultural theory} {Explaining the dialogic processes of teaching and learning: {The} value and potential of sociocultural theory}.{\BBCQ}
\newblock
\APACjournalVolNumPages{Learning, Culture and Social Interaction}{1}{1}{12--21,}
\newblock
\begin{APACrefDOI} \doi{10.1016/j.lcsi.2012.03.001} \end{APACrefDOI}
\newblock
\begin{APACrefURL} [{2025-04-14}]{https://www.sciencedirect.com/science/article/pii/S2210656112000049} \end{APACrefURL}
\newblock

\newblock

\PrintBackRefs{\CurrentBib}

\bibitem [\protect \citeauthoryear {%
Mercer%
, Lyn%
\BCBL {}\ \BBA {} Staarman%
}{%
Mercer%
\ \protect \BOthers {.}}{%
{\protect \APACyear {2009}}%
}]{%
mercer_dialogic_2009}
\APACinsertmetastar {%
mercer_dialogic_2009}%
\begin{APACrefauthors}%
Mercer, N.%
, Lyn, D.%
\BCBL {} Staarman, J.%
\end{APACrefauthors}%
\unskip\
\newblock
\APACrefYearMonthDay{2009}{{\APACmonth{07}}}{}.
\newblock
{\BBOQ}\APACrefatitle {Dialogic teaching in the primary science classroom} {Dialogic teaching in the primary science classroom}.{\BBCQ}
\newblock
\APACjournalVolNumPages{Language and Education}{23}{4}{353--369,}
\newblock
\begin{APACrefDOI} \doi{10.1080/09500780902954273} \end{APACrefDOI}
\newblock
\begin{APACrefURL} [{2025-04-14}]{https://doi.org/10.1080/09500780902954273} \end{APACrefURL}
\newblock
\APACrefnote{Publisher: Routledge \_eprint: https://doi.org/10.1080/09500780902954273}
\newblock

\newblock

\PrintBackRefs{\CurrentBib}

\bibitem [\protect \citeauthoryear {%
Piaget%
}{%
Piaget%
}{%
{\protect \APACyear {1964}}%
}]{%
piaget_part_1964}
\APACinsertmetastar {%
piaget_part_1964}%
\begin{APACrefauthors}%
Piaget, J.%
\end{APACrefauthors}%
\unskip\
\newblock
\APACrefYearMonthDay{1964}{}{}.
\newblock
{\BBOQ}\APACrefatitle {Part {I}: {Cognitive} development in children: {Piaget} development and learning} {Part {I}: {Cognitive} development in children: {Piaget} development and learning}.{\BBCQ}
\newblock
\APACjournalVolNumPages{Journal of Research in Science Teaching}{2}{3}{176--186,}
\newblock
\begin{APACrefDOI} \doi{10.1002/tea.3660020306} \end{APACrefDOI}
\newblock
\begin{APACrefURL} [{2025-04-14}]{https://onlinelibrary.wiley.com/doi/abs/10.1002/tea.3660020306} \end{APACrefURL}
\newblock
\APACrefnote{\_eprint: https://onlinelibrary.wiley.com/doi/pdf/10.1002/tea.3660020306}
\newblock

\newblock

\PrintBackRefs{\CurrentBib}

\bibitem [\protect \citeauthoryear {%
Sinha%
}{%
Sinha%
}{%
{\protect \APACyear {2025}}%
}]{%
sinha2025beyond}
\APACinsertmetastar {%
sinha2025beyond}%
\begin{APACrefauthors}%
Sinha, T.%
\end{APACrefauthors}%
\unskip\
\newblock
\APACrefYearMonthDay{2025}{}{}.
\newblock
{\BBOQ}\APACrefatitle {Beyond good {AI}: {The} need for sound learning theories in {AIED}} {Beyond good {AI}: {The} need for sound learning theories in {AIED}}.{\BBCQ}
\newblock
\APACjournalVolNumPages{Technology, Knowledge and Learning}{}{}{,}
\newblock
\begin{APACrefDOI} \doi{10.1007/s10758-025-09843-9} \end{APACrefDOI}
\newblock

\newblock

\PrintBackRefs{\CurrentBib}

\bibitem [\protect \citeauthoryear {%
Sætra%
}{%
Sætra%
}{%
{\protect \APACyear {2025}}%
}]{%
satra2025scaffolding}
\APACinsertmetastar {%
satra2025scaffolding}%
\begin{APACrefauthors}%
Sætra, H.S.%
\end{APACrefauthors}%
\unskip\
\newblock
\APACrefYearMonthDay{2025}{}{}.
\newblock
{\BBOQ}\APACrefatitle {Scaffolding human champions: {AI} as a more competent other} {Scaffolding human champions: {AI} as a more competent other}.{\BBCQ}
\newblock
\APACjournalVolNumPages{Human Arenas}{8}{}{56--78,}
\newblock
\begin{APACrefDOI} \doi{10.1007/s42087-022-00304-8} \end{APACrefDOI}
\newblock

\newblock

\PrintBackRefs{\CurrentBib}

\bibitem [\protect \citeauthoryear {%
Tang%
\ \protect \BOthers {.}}{%
Tang%
\ \protect \BOthers {.}}{%
{\protect \APACyear {2024}}%
}]{%
tang_dialogic_2024}
\APACinsertmetastar {%
tang_dialogic_2024}%
\begin{APACrefauthors}%
Tang, K\BHBI S.%
, Cooper, G.%
, Rappa, N.%
, Cooper, M.%
, Sims, C.%
\BCBL {} Nonis, K.%
\end{APACrefauthors}%
\unskip\
\newblock
\APACrefYearMonthDay{2024}{{\APACmonth{02}}}{}.
\newblock
\APACrefbtitle {A dialogic approach to transform teaching, learning \&assessment with generative {AI} in secondary education:a proof of concept} {A dialogic approach to transform teaching, learning \&assessment with generative {AI} in secondary education:a proof of concept}\ [{SSRN} {Scholarly} {Paper}].
\newblock
\APACaddressPublisher{Rochester, NY}{Social Science Research Network}.
\newblock
\begin{APACrefURL} [{2025-04-14}]{https://papers.ssrn.com/abstract=4722537} \end{APACrefURL}
\PrintBackRefs{\CurrentBib}

\bibitem [\protect \citeauthoryear {%
Thomas%
}{%
Thomas%
}{%
{\protect \APACyear {1994}}%
}]{%
thomas_conversational_1994}
\APACinsertmetastar {%
thomas_conversational_1994}%
\begin{APACrefauthors}%
Thomas, A.%
\end{APACrefauthors}%
\unskip\
\newblock
\APACrefYearMonthDay{1994}{}{}.
\newblock
{\BBOQ}\APACrefatitle {Conversational {Learning}} {Conversational {Learning}}.{\BBCQ}
\newblock
\APACjournalVolNumPages{Oxford Review of Education}{20}{1}{131--142,}
\newblock
\begin{APACrefURL} [{2025-04-09}]{https://www.jstor.org/stable/1050899} \end{APACrefURL}
\newblock
\APACrefnote{Publisher: Taylor \& Francis, Ltd.}
\newblock

\newblock

\PrintBackRefs{\CurrentBib}

\bibitem [\protect \citeauthoryear {%
Tran%
\ \protect \BOthers {.}}{%
Tran%
\ \protect \BOthers {.}}{%
{\protect \APACyear {2025}}%
}]{%
tran_multi-agent_2025}
\APACinsertmetastar {%
tran_multi-agent_2025}%
\begin{APACrefauthors}%
Tran, K\BHBI T.%
, Dao, D.%
, Nguyen, M\BHBI D.%
, Pham, Q\BHBI V.%
, O'Sullivan, B.%
\BCBL {} Nguyen, H.D.%
\end{APACrefauthors}%
\unskip\
\newblock
\APACrefYearMonthDay{2025}{{\APACmonth{01}}}{}.
\newblock
\APACrefbtitle {Multi-{Agent} {Collaboration} {Mechanisms}: {A} {Survey} of {LLMs}.} {Multi-{Agent} {Collaboration} {Mechanisms}: {A} {Survey} of {LLMs}.}
\newblock
\APACaddressPublisher{}{arXiv}.
\newblock
\begin{APACrefURL} [{2025-04-14}]{http://arxiv.org/abs/2501.06322} \end{APACrefURL}
\newblock
\APACrefnote{arXiv:2501.06322 [cs]}
\PrintBackRefs{\CurrentBib}

\bibitem [\protect \citeauthoryear {%
Vygotsky%
}{%
Vygotsky%
}{%
{\protect \APACyear {1978}}%
}]{%
vygotsky_mind_1978}
\APACinsertmetastar {%
vygotsky_mind_1978}%
\begin{APACrefauthors}%
Vygotsky, L.S.%
\end{APACrefauthors}%
\unskip\
\newblock
\APACrefYear{1978}.
\newblock
\APACrefbtitle {Mind in {Society}: {The} {Development} of {Higher} {Psychological} {Processes}} {Mind in {Society}: {The} {Development} of {Higher} {Psychological} {Processes}}.
\newblock
\APACaddressPublisher{}{Harvard University Press}.
\PrintBackRefs{\CurrentBib}

\bibitem [\protect \citeauthoryear {%
Wegerif%
}{%
Wegerif%
}{%
{\protect \APACyear {2013}}%
}]{%
wegerif2013dialogic}
\APACinsertmetastar {%
wegerif2013dialogic}%
\begin{APACrefauthors}%
Wegerif, R.%
\end{APACrefauthors}%
\unskip\
\newblock
\APACrefYear{2013}.
\newblock
\APACrefbtitle {Dialogic: {Education} for the internet age} {Dialogic: {Education} for the internet age}.
\newblock
\APACaddressPublisher{}{Routledge}.
\PrintBackRefs{\CurrentBib}

\bibitem [\protect \citeauthoryear {%
Ying%
\ \protect \BOthers {.}}{%
Ying%
\ \protect \BOthers {.}}{%
{\protect \APACyear {2019}}%
}]{%
ying_their_2019}
\APACinsertmetastar {%
ying_their_2019}%
\begin{APACrefauthors}%
Ying, K.M.%
, Pezzullo, L.G.%
, Ahmed, M.%
, Crompton, K.%
, Blanchard, J.%
\BCBL {} Boyer, K.E.%
\end{APACrefauthors}%
\unskip\
\newblock
\APACrefYearMonthDay{2019}{{\APACmonth{02}}}{}.
\newblock
{\BBOQ}\APACrefatitle {In {Their} {Own} {Words}: {Gender} {Differences} in {Student} {Perceptions} of {Pair} {Programming}} {In {Their} {Own} {Words}: {Gender} {Differences} in {Student} {Perceptions} of {Pair} {Programming}}.{\BBCQ}
\newblock
 \APACrefbtitle {Proceedings of the 50th {ACM} {Technical} {Symposium} on {Computer} {Science} {Education}} {Proceedings of the 50th {ACM} {Technical} {Symposium} on {Computer} {Science} {Education}}\ (\BPGS\ 1053--1059).
\newblock
\APACaddressPublisher{New York, NY, USA}{Association for Computing Machinery}.
\newblock
\begin{APACrefURL} [{2025-04-14}]{https://dl.acm.org/doi/10.1145/3287324.3287380} \end{APACrefURL}
\PrintBackRefs{\CurrentBib}

\bibitem [\protect \citeauthoryear {%
Yusuf%
, Money%
\BCBL {}\ \BBA {} Daylamani-Zad%
}{%
Yusuf%
\ \protect \BOthers {.}}{%
{\protect \APACyear {2025}}%
}]{%
yusuf_pedagogical_2025}
\APACinsertmetastar {%
yusuf_pedagogical_2025}%
\begin{APACrefauthors}%
Yusuf, H.%
, Money, A.%
\BCBL {} Daylamani-Zad, D.%
\end{APACrefauthors}%
\unskip\
\newblock
\APACrefYearMonthDay{2025}{{\APACmonth{01}}}{}.
\newblock
{\BBOQ}\APACrefatitle {Pedagogical {AI} Conversational agents in higher education: a conceptual framework and survey of the state of the art} {Pedagogical {AI} conversational agents in higher education: a conceptual framework and survey of the state of the art}.{\BBCQ}
\newblock
\APACjournalVolNumPages{Educational technology research and development}{}{}{,}
\newblock
\begin{APACrefDOI} \doi{10.1007/s11423-025-10447-4} \end{APACrefDOI}
\newblock
\begin{APACrefURL} [{2025-04-09}]{https://doi.org/10.1007/s11423-025-10447-4} \end{APACrefURL}
\newblock

\newblock

\PrintBackRefs{\CurrentBib}

\bibitem [\protect \citeauthoryear {%
Zhang%
, Zou%
\BCBL {}\ \BBA {} Cheng%
}{%
Zhang%
\ \protect \BOthers {.}}{%
{\protect \APACyear {2024}}%
}]{%
zhang2024review}
\APACinsertmetastar {%
zhang2024review}%
\begin{APACrefauthors}%
Zhang, R.%
, Zou, D.%
\BCBL {} Cheng, G.%
\end{APACrefauthors}%
\unskip\
\newblock
\APACrefYearMonthDay{2024}{}{}.
\newblock
{\BBOQ}\APACrefatitle {A review of chatbot-assisted learning: {Pedagogical} approaches, implementations, and future directions} {A review of chatbot-assisted learning: {Pedagogical} approaches, implementations, and future directions}.{\BBCQ}
\newblock
\APACjournalVolNumPages{Interactive Learning Environments}{32}{8}{4529--4557,}
\newblock
\begin{APACrefDOI} \doi{10.1080/10494820.2023.2202704} \end{APACrefDOI}
\newblock

\newblock

\PrintBackRefs{\CurrentBib}

\bibitem [\protect \citeauthoryear {%
Zimmerman%
}{%
Zimmerman%
}{%
{\protect \APACyear {2002}}%
}]{%
zimmerman2002becoming}
\APACinsertmetastar {%
zimmerman2002becoming}%
\begin{APACrefauthors}%
Zimmerman, B.J.%
\end{APACrefauthors}%
\unskip\
\newblock
\APACrefYearMonthDay{2002}{}{}.
\newblock
{\BBOQ}\APACrefatitle {Becoming a self-regulated learner: {An} overview} {Becoming a self-regulated learner: {An} overview}.{\BBCQ}
\newblock
\APACjournalVolNumPages{Theory into practice}{41}{2}{64--70,}
\newblock
\APACrefnote{Publisher: Taylor \& Francis}
\newblock

\newblock

\PrintBackRefs{\CurrentBib}

\end{thebibliography}
\end{document}